%% file: neurips_2026.tex
\documentclass{article}

 \usepackage[main, final]{neurips_2026}

\usepackage[utf8]{inputenc} 
\usepackage[T1]{fontenc}    
\usepackage{url}            
\usepackage{booktabs}       
\usepackage{amsfonts}       
\usepackage{nicefrac}       
\usepackage{microtype}      
\usepackage{xcolor}         

\usepackage{multirow}
\usepackage{array}
\usepackage{graphicx} 
\graphicspath{{./images/}}
\usepackage{subcaption}
\usepackage{caption}
\usepackage{bm}
\usepackage{tabularx}
\usepackage{wrapfig}
\usepackage{enumitem}
\usepackage{seqsplit}
\usepackage{relsize}
\newcommand{\gray}[1]{\textcolor{gray}{#1}}
\newcolumntype{Y}{>{\centering\arraybackslash}X}
\usepackage{marvosym}
\usepackage{amsmath}
\usepackage{amssymb}
\usepackage{mathtools}
\usepackage{amsthm}
\newtheorem{theorem}{Theorem}  

\definecolor{mycitecolor}{HTML}{3c6db0}
\definecolor{mylinkcolor}{HTML}{bc3c53} 
\usepackage[colorlinks=true, citecolor=mycitecolor, linkcolor=mylinkcolor,urlcolor=magenta]{hyperref}

\title{Efficient Dataset Distillation for Pre-Trained Self-Supervised Models via Statistical Flow Matching}

\author{Qianxin Xia\textsuperscript{1,2*} \quad
Jiawei Du\textsuperscript{2 \Letter}\quad
Yuhan Zhang\textsuperscript{3}\quad 
Xin Zhang\textsuperscript{2} \\
\textbf{Xuewan He\textsuperscript{1}}\quad
\textbf{Wenbo Jiang\textsuperscript{1}}\quad
\textbf{Jielei Wang\textsuperscript{1}}\quad
\textbf{Tao Luo\textsuperscript{2}}\quad
\textbf{Guoming Lu\textsuperscript{1 \Letter}}\\
\textsuperscript{1}{\small University of Electronic Science and Technology of China, China} \\
\hspace{-0.6em}\textsuperscript{2}{\small Institute of Advanced Intelligence and Computing, Agency for Science, Technology and Research, Singapore}\\
\textsuperscript{3}{\small Sichuan University, China} \\
\texttt{\small \{xqx023, hershel\}@std.uestc.edu.cn} \quad \texttt{\small \{wenbo\_jiang, jlwang, lugm\}@uestc.edu.cn} \\
\texttt{\small \{dujw, zhangx7, luo\_tao\}@a-star.edu.sg} \quad \texttt{\small zyhyuhan@stu.scu.edu.cn}
}

\begin{document}

\maketitle

\begingroup
\renewcommand{\thefootnote}{}
\footnotetext{\hspace{-1.8em}\Letter\ Corresponding author.
\\ \hspace*{0.21em}* Work completed during internship at IAIC, A*STAR.}
\endgroup

\begin{abstract}
 Dataset distillation seeks to synthesize a compact surrogate dataset that enables performance comparable to training on the original dataset for downstream tasks. For the scenario where pre-trained self-supervised models serve as priors, traditional \textit{Linear Gradient Matching} optimizes synthetic images by encouraging them to mimic the gradient updates induced by real images on the linear probe. However, this batch-level formulation requires loading thousands of real images and applying multiple differentiable augmentations to synthetic images at each distillation step, leading to substantial computational and memory overheads. In this paper, we revisit the linear gradient and theoretically derive that it is essentially a \textit{local} relative distribution directed from target class centers toward non-target class centers, which we term ``flow”. This property causes suboptimality and instability, often necessitating expensive multiple augmentations to compensate. To address this, we introduce \textit{Statistical Flow Matching}, an optimal, stable, and efficient supervised learning framework that optimizes synthetic images by aligning \textit{global} statistical flows in the original data. Our approach loads raw statistics only once and performs a single augmentation pass on the synthetic data, achieving performance comparable to or better than the state-of-the-art method with 10× less GPU memory usage and 4× faster distillation time. Moreover, increasing the number of augmentations for our method yields further performance gains while incurring lower additional cost. Our code is publicly available at \url{https://github.com/einsteinxia/SFM}.
\end{abstract}

\section{Introduction}
In machine learning, data serves as the “fuel” that drives improvements in model capabilities. As datasets grow larger and more diverse, models can learn richer patterns and achieve better performance. However, this reliance on data also introduces significant practical overheads, since large-scale dataset acquisition, management, transmission, and processing require substantial storage, communication and compute capacity \cite{lei2023comprehensive}. To reconcile high performance with prohibitive resource costs, \textit{Dataset Distillation} (DD) \cite{wang2018dataset} has been proposed to compress a cumbersome dataset into a compact surrogate that yields training efficacy comparable to using the full dataset. This enables more efficient model training and benefits downstream tasks such as continual learning \cite{yang2023efficient, ohasymmetric} and neural architecture search \cite{medvedev2021learning, ding2024calibrated}. On the other hand, the application of DD has been extended by numerous works to a range of research fields, e.g., privacy protection \cite{yang2026sneakdoor, zheng2025improving} and federated learning \cite{yan2025fedvck, jia2024unlocking}.

Several recent methods \cite{zhao2023dataset, cazenavette2022dataset, yin2023squeeze, su2024d} and their extensions \cite{wang2025dataset, du2023minimizing, shao2024generalized, xia2026diver} have advanced the ambitious goal of learning high-quality models from synthetic images. However, they typically require distilling a relatively large Image-Per-Class (IPC) and training models from scratch to achieve merely mediocre or modest performance, both of which are inefficient and limit practical training and deployment in specific scenarios, such as resource-constrained edge environments. Consequently, there is an urgent need to investigate a more efficient, practice-ready scheme to break this impasse.

Fortunately, computer vision has increasingly shifted toward exploiting representations learned by large-scale pre-trained self-supervised vision models (e.g., DINO \cite{oquab2023dinov2}, CLIP \cite{radford2021learning}, EVA \cite{fang2024eva}, and MoCo \cite{chen2021empirical}) for downstream tasks. In practice, such models are either lightly fine-tuned on the target task or used as frozen backbone encoders to provide transferable features for lightweight task-specific heads. With these strong priors, competitive models can often be obtained rapidly using only a small amount of data \cite{simeoni2025dinov3}. Therefore, we argue that developing efficient data synthesis and training strategies tailored to such pre-trained self-supervised vision models is a key prerequisite for advancing DD beyond a “toy” setting toward real-world applicability.

\input{figures/motivation}

The pioneering work, Linear Gradient Matching (LGM) \cite{cazenavette2025lgm}, first forwards both real images and noise-initialized synthetic images through the frozen backbone of the pre-trained self-supervised model, and then optimizes the synthetic images so that they induce gradient updates on the backbone-attached linear classification head that are similar to those induced by real images. On the one hand, this batch-level formulation requires loading numerous real images at each distillation step. On the other hand, each synthetic image undergoes multiple rounds of differentiable augmentations, so the \textit{autograd} computation graphs for all views and their intermediate activations must be retained to enable backpropagation, which incurs significant computational overhead and GPU memory usage.

To address these issues, we delve into the mechanism of LGM. We first theoretically reformulate the linear gradient into an analytical form of local relative distribution, which we call \textbf{flow}. The flow is directed from the \textbf{target class}'s distribution center toward that of the \textbf{non-target class}. Subsequently, we reveal some key limitations (e.g., suboptimality, instability, and inefficiency) of this local matching paradigm and propose a novel supervised learning framework called \textbf{S}tatistical \textbf{F}low \textbf{M}atching \textbf{(SFM)}. SFM first statistically summarizes the global distribution center of the original dataset just once to form a constant flow, which remains fixed throughout the optimization. It then induces the synthetic flow of the distilled images to align with the statistical flow, achieving optimal, stable, and efficient synthesis.

As shown in Figure \ref{fig:motivation}, SFM requires training only a linear classifier with IPC=1, taking just seconds, yet achieves 80\%$\sim$90\% validation accuracy across multiple self-supervised models. Compared to LGM with ten augmentations, our SFM with just a single augmentation achieves comparable distillation performance and superior generalization performance, while reducing GPU memory usage by 10× (16 GB vs. 165 GB) and shortening the distillation time by 4× (18 min vs. 80 min). Moreover, applying more augmentations to SFM yields further gains, confirming the favorable scalability. In practice, we can also adjust the trade-off between resources and performance according to requirements. Notably, since SFM does not require frequently loading large batches of real images from the dataloader or performing additional gradient computations at each distillation step, our distillation time is significantly lower than that of LGM under the same augmentation settings.

Overall, our contributions are summarized as follows:

\begin{itemize}
\item We theoretically identify linear gradient as a local relative distribution, thereby establishing for the first time the concept of ``flow” for dataset distillation.
\item We propose statistical flow matching, a novel framework for global supervised learning that ensures optimality, stability, and efficiency in the process of data synthesis.
\item Extensive experiments demonstrate that our method consistently outperforms the state-of-the-art linear gradient matching in both performance and efficiency.

\end{itemize}

\section{Related Work}
\textbf{Dataset Distillation.} As the scale of datasets in machine learning continues to grow, efficient storage and learning paradigms have attracted the attention of researchers from various fields. As a result, they have shifted from the initial coreset approach \cite{chen2012super, wei2015submodularity}, which directly selects representative samples from the raw dataset, to dataset distillation, which generates abstract or non-raw images containing insightful information from the original dataset. This has led to a range of modes, including the introduction of consistency loss that matches distribution \cite{li2025diversity}, gradient \cite{zhao2020dataset}, trajectory \cite{zhong2025towards}, statistics \cite{shao2024generalized},  and other parameter optimization \cite{bao2025dataset}, generation-based approaches \cite{cazenavette2023generalizing, chan2025mgd}. However, under low-IPC settings, these methods do not perform particularly well when training models from scratch. Even under high IPC configurations, they exhibit only modest to mediocre performance while resource requirements grow proportionally with IPC. LGM \cite{cazenavette2025lgm} is the first to incorporate pre-trained model priors into DD to bridge the above performance gap, but at a prohibitive cost.


\textbf{Self-Supervised Learning.} Due to the expensive and time-consuming nature of collecting labeled data on a large scale in reality, self-supervised learning has become a representative method for pre-training networks, allowing models to learn transferable representations from abundant unlabeled data for subsequent downstream tasks. Widely adopted self-supervised vision models mainly include CLIP \cite{radford2021learning}, DINO \cite{oquab2023dinov2}, EVA \cite{fang2024eva}, and MoCo \cite{chen2021empirical}.

In this work, we consider an extreme setting where we distill only \textbf{one image per class} from pre-trained self-supervised vision models, and for downstream tasks we train only \textbf{a linear classifier}. In addition, we mainly address the limitations imposed by LGM, including suboptimality, instability, and inefficiency. The satisfactory results have indeed unlocked the potential of this paradigm.

\section{Method}
In this section, we revisit linear gradient matching in pre-trained self-supervised vision models and deeply explore its mechanism. We reveal its essence and the limitations it brings. Based on this, we propose statistical flow matching, thereby improving distillation efficiency and performance. An overview of our SFM and the comparison to LGM are presented in Figure \ref{fig:overview}.

\subsection{Rethinking Linear Gradient Matching}

Formally, we consider a frozen pre-trained self-supervised model or feature extractor $\bm{\phi}$ that maps images $\bm{x}$ into a latent representation space $\bm{\phi}(\bm{x})^{F \times 1}$, where $F$ is the output dimension of $\bm{\phi}$.  Given an original large-scale dataset $\mathcal{O}={\{({\bm{x}}_{i}^{o}, {\bm{y}}_{i}^{o})\}}_{i=1}^{|\mathcal{O}|}$, our objective is to construct a compact synthetic dataset $\mathcal{S}=\{({\bm{x}}_{i}^{s}, {\bm{y}}_{i}^{s})\}_{i=1}^{|\mathcal{S}|}$, which enables training a linear classification head on top of $\bm{\phi}$ with performance on par with that trained on the full real dataset. where $|\mathcal{S}|  \ll  |\mathcal{O}|$ and ${\bm{x}}_{i}^{o}, {\bm{x}}_{i}^{s}\sim {P}(\bm{x}^{o}),{P}({\bm{x}}^{s})$ are the $i$-th original and synthetic images with the corresponding one-hot labels ${\bm{y}}_{i}^{o}, {\bm{y}}_{i}^{s} \in \mathcal{Y} ^{C \times 1}$. $C$ is the number of classes.

To achieve this, LGM samples a \textbf{random} linear classifier matrix \( \bm{W}\sim \mathcal{N}(0, {0.01  ^{\text{2}}}) \in \mathbb{R}^{C \times F} \) at each distillation step to enforce that training on the synthetic data mimics the gradient updates from training on real data under diverse initialization. After the original and synthetic images pass forward through $\bm{\phi}$ and $\bm{W}$, the gradient of the cross-entropy loss $\bm{\ell}$ with respect to $\bm{W}$ is:
\begin{equation}
\frac{\partial \bm{\ell}}{\partial \bm{W}}= \frac{1}{B} \sum_{i=1}^{B}
(\bm{p}_i - \bm{y}_i){\bm{\phi}(\bm{x}_i})^{\top},
\label{eq:batch_grad}
\end{equation}
where ${B}=aC$ is the batch size under the setting of IPC=1, $a$ is the number of augmentation for each class.  $\bm{p}_i=[{p}_{i1},{p}_{i2},\cdots ,{p}_{ic},\cdots ,{p}_{iC}]^{\top}$ is the probability distribution of the $i$-th sample:
\begin{equation}
{p}_{ic}=\text{softmax}({z}_{ic})=\frac{\exp\bigl({{z}_{ic}})}{\textstyle\sum_{j=1}^{C}\exp\bigl({{z}_{ij}})},   
\label{eq:softmax}
\end{equation}
where $\bm{z}_i=[{z}_{i1},{z}_{i2},\cdots ,{z}_{ic},\cdots ,{z}_{iC}]^{\top}=\bm{W}\bm{\phi}({\bm{x}}_{i})$ is the logits of the $i$-th sample.

Then, the LGM loss can be defined as:
\begin{equation}
\mathcal{L}_\mathit{lgm} = 1-cos[flat(\frac{\partial \bm{\ell}^o}{\partial \bm{W}}), flat(\frac{\partial \bm{\ell}^s}{\partial \bm{W}})],
\label{eq:lgm}
\end{equation}
where $flat$ represents the $flatten$ operation. $ \bm{\ell}^o$ and $ \bm{\ell}^s$ represent the losses on the original image ``o” and the synthetic image ``s”, respectively. $\mathcal{L}_{lgm}$ is subsequently backpropagated through the computational graph, which includes the inner gradient step, feature extractor, and the linear classifier, to obtain $\partial \mathcal{L}_{lgm}/{\partial \bm{x}^{s}}$ and update the synthetic images.

For Equation \ref{eq:batch_grad}, given the determinism of the labels $\bm{y}$ and the feature extractor $\bm{\phi}$, the update of the synthetic data depends entirely on the probability distribution  $\bm{p}$, which is in turn governed exclusively by the dynamic classifier's linear matrix $\bm{W}$. 

\input{figures/overview}

\begin{theorem}[Exchangeability]
In a multiclass classification using the softmax function. Let the number of classes be $C$, and the weight vectors $\bm{W}_{c}\in\mathbb{R}^{F \times 1}$ for each class $c$ are independently initialized from a Gaussian distribution with mean zero  $\bm{W}_{c}\sim \mathcal{N}(0, { \sigma_{c}^{\text{2}}})$. For sample $i$, the predicted class probabilities ${p}_{ic}$ for each class $c$ are exchangeable and uniformly distributed across the classes:
\begin{equation*}
\mathbb{E}[{p}_{i1}] = \cdots = \mathbb{E}[{p}_{ic}] = \cdots = \mathbb{E}[{p}_{iC}] = 1/C.
\label{eq:exchangeable}
\end{equation*}
\label{th:exchangeability}
\vspace{-10pt}
\end{theorem}
\begin{theorem}[LogNormal Distribution]
Let $Y \sim \mathcal{N}(\mu, \sigma^2)$ be a normally distributed random variable, and define $X = e^{Y}$.
Then $X$ follows a lognormal distribution, denoted $X \sim \mathcal{LN}(\mu, \sigma^2)$, with:
\begin{align*}
\mathbb{E}[X] =\exp\bigl(\mu + \sigma^{2}/{2}), \quad \operatorname{Var}[X] = \exp\bigl(2\mu + \sigma^{2}\bigr)\bigl[\exp(\sigma^{2}) - 1\bigr].
\end{align*}
\label{th:lognormal}
\end{theorem}
\vspace{-10pt}
According to \textbf{Theorem \ref{th:exchangeability}}, the expectation of Equation \ref{eq:batch_grad} is:
\begin{equation}
\begin{aligned}
\mathbb{E}[\frac{\partial \bm{\ell}}{\partial \bm{W}}] = \frac{1}{B} \sum_{i=1}^{B}
(\mathbb{E}[\bm{p}_i] - \bm{y}_i){\bm{\phi}(\bm{x}_i})^{\top} 
= \frac{1}{B} \sum_{i=1}^{B}
(\frac{1}{C}\bm{1} - \bm{y}_i){\bm{\phi}(\bm{x}_i})^{\top}.
\end{aligned}
\label{eq:batch_grad_exp}
\end{equation}
We are aware that the $c$-th row of linear classifier matrix  \( \bm{W}_{c}\sim \mathcal{N}(0, {0.01 ^{\text{2}}}) \), ${z}_{ic}=\bm{W}_c\bm{\phi}({\bm{x}}_{i})\sim \mathcal{N}(0, {\sigma^2})$ ($\sigma = {0.01 \cdot ||\bm{\phi}({\bm{x}}_{i}||}$). A LayerNorm operation is typically applied at the end of self-supervised models, which ensures that $||\bm{\phi}({\bm{x}}_{i}||=O(1)$ and $\sigma$ has a bounded and small value. 

According to \textbf{Theorem \ref{th:lognormal}}, the expectation and variance of $\exp\bigl({{z}_{ic}})$ can be expresses as:
\begin{equation}
\begin{aligned}
\mathbb{E}[\exp({{z}_{ic}})]=\exp\bigl({\sigma ^2/2}), \quad
\operatorname{Var}[\exp({{z}_{ic}})]=\exp\bigl({\sigma ^2})\cdot[\exp\bigl({\sigma ^2})-1],
\end{aligned}
\label{eq:log_normal}
\end{equation}
when $C$ is relatively large (e.g., 100), according to the law of large numbers  $ \mathbb{E}[\exp({{z}_{ic}})]\approx \textstyle\sum_{j=1}^{C}\exp({{z}_{ij}}) /C$, the variance of ${p}_{ic}$ in Equation \ref{eq:softmax} can be expressed as:
\begin{equation}
\begin{aligned}
\operatorname{Var}[{p}_{ic}]&= \operatorname{Var}[\frac{\exp\bigl({{z}_{ic}})}{C\cdot \exp\bigl({\sigma ^2/2})}]=\frac{\operatorname{Var}[\exp\bigl({{z}_{ic}})]}{{C}^{2}\cdot\exp\bigl({\sigma ^2})} \\
&=\frac{\exp\bigl({\sigma ^2})\cdot[\exp\bigl({\sigma ^2})-1]}{{C}^{2} \cdot \exp\bigl({\sigma ^2})}=\frac{\exp\bigl({\sigma ^2})-1}{C^2},
\end{aligned}
\label{eq:variance}
\end{equation}

since $\sigma$ is very small, we can see that $\operatorname{Var}[{p}_{ic}]\rightarrow0$. According to \textbf{Theorem \ref{th:exchangeability}} : $\mathbb{E}[{p}_{ic}] = 1/C$, ${p}_{ic}\rightarrow1/C$ and its slight fluctuation is determined solely by \( \bm{W}\). 

\input{tables/analytic}
To investigate the impact of \( \bm{W}\), we establish three baselines: (1) the raw LGM with \textbf{random} \( \bm{W}\) sampled at every distillation step, (2) \textbf{fixed} \( \bm{W}\) over the whole distillation process, and (3) an \textbf{analytical} form in Equation \ref{eq:batch_grad_exp} that is independent of \( \bm{W}\). As shown in Table \ref{tab:analytic}, all modes under various models achieve similar performance, which indicates that \( \bm{W}\) has no significant impact on performance. This primarily stems from the fact that ${p}_{ic}\rightarrow1/C$ is weakly correlated with \( \bm{W}\). Although randomly sampling \( \bm{W}\) intends to enhance generalization by aligning gradients under different initial conditions, both our theoretical analysis and experimental results indicate that this constitutes a redundant constraint.

Therefore, the gradient of the loss $\bm{\ell}$ with respect to  \( \bm{W}_{c}\) for each class $c$ degenerates entirely into the analytical form shown in Equation \ref{eq:batch_grad_exp}:
\begin{equation}
\small
\begin{aligned}
\frac{\partial \bm{\ell}}{\partial \bm{W}_c} 
&= \frac{1}{B} \sum_{i=1}^{B}
\frac{1}{C}{\bm{\phi}(\bm{x}_i|_{\bm{y}_{ic}=0}})^{\top} - \frac{C-1}{C}{\bm{\phi}(\bm{x}_i|_{\bm{y}_{ic}=1}})^{\top}  \\
&=\frac{1}{aC}\cdot \frac{a(C-1)}{C} \cdot
[\frac{\displaystyle\sum_{i=1}^{a(C-1)}{{\phi}({x}_i|_{{y}_{ic}=0}})^{\top}}{a(C-1)}- \frac{\displaystyle\sum_{i=1}^{a}{{\phi}({x}_i|_{{y}_{ic}=1}})^{\top}}{a}]  \\[0.8\baselineskip]
&=\frac{C-1}{C^2} [{\overline{\phi}(\bm{x}|_{\bm{y}_{c}=0}})^{\top}- {\overline{\phi}(\bm{x}|_{\bm{y}_{c}=1}})^{\top}],
\label{eq:relative_distribution}
\end{aligned}
\end{equation}
where the overline ``$-$” denotes the mean. This effectively models a pattern of relative distribution. We refer to it as \textbf{flow} and the corresponding matching strategy is called flow matching. The flow originates from the distribution center of \textbf{target class} ${\overline{\phi}({x}|_{{y}_{c}=1}})$ and terminates at the distribution center of \textbf{non-target class} ${\overline{\phi}({x}|_{{y}_{c}=0}})$. Notably, our terminology is somewhat similar to that of flow-based generative models \cite{lipman2022flow, liu2022flow}, but fundamentally different. In generative models, a flow typically refers to a mapping from a Gaussian distribution to the data distribution, whereas our flow denotes a vector representation pointing from the target class center to the non-target class center. 


\subsection{Statistical Flow Matching}

In Equation \ref{eq:relative_distribution}, we derive the analytical form of the linear gradient and show that, given a fixed number of classes $C$, it depends solely on the distribution centers of the data. For the original dataset, the matching strategy is invariably performed at the batch level, which results in some limitations:

\textbf{Suboptimality.} The local target and non-target class centers inadequately represent the full data distribution within a batch, leading to biased optimization that deviates from the global optimum. This limitation becomes more pronounced when the batch size is small.

\textbf{Instability.} The centers or flows vary dynamically across distillation steps or batches, which causes optimization instability that induces artifacts in Figure \ref{fig:syn_images} (left) and thereby degrades performance. These also explain why LGM implicitly adopts a large batch size on original data and multiple augmentations on synthetic data, which proves to be generally effective in capturing global distribution. 

\textbf{Inefficiency.} Employing a large batch size and extensive augmentations can alleviate the negative effects of the above two limitations to some extent, but they  incur substantial demands on memory and computational resources. This stems from the necessity to load large volumes of original images and maintain intricate computational graphs for gradient computation across numerous augmented synthetic images, rendering the process highly inefficient.

Consequently, we propose a simple but comprehensive method to simultaneously address the aforementioned three issues. We precompute class-wise statistical centers ${\phi}^{*}(\bm{x})$ from the original dataset ``o”, thereby constructing a constant statistical flow $\bm{\mathcal{F}}^{*}=[\mathcal{F}_1^{*}, \cdots, \mathcal{F}_c^{*}, \cdots, \mathcal{F}_C^{*}]$, where: 
\begin{equation}
\mathcal{F}_c^{*}={\phi}^{*}(\bm{x}^{o}|_{\bm{y}^{o}_{c}=0})^{\top}-{\phi}^{*}(\bm{x}^{o}|_{\bm{y}^{o}_{c}=1})^{\top}.
\label{eq:global_flow}
\end{equation}
The statistical flow remains fixed during the optimization process, enabling synthetic data to learn a global objective in an optimal, stable, and efficient manner. Our meta loss is designed as the cosine distance between the synthetic flow $\bm{\mathcal{F}}^{s}$ and the statistical flow $\bm{\mathcal{F}}^{*}$:
\begin{equation}
\mathcal{L}_\mathit{sfm} = 1-cos[flat(\bm{\mathcal{F}}^{*}), flat(\bm{\mathcal{F}}^{s})].
\label{eq:sfm}
\end{equation}
Operating within a \textbf{supervised learning} paradigm, our method can stably capture the latent structure in the original data. This achieves efficient optimization without mandating the loading of numerous images from the original data and the execution of multiple augmentations on synthetic data.

\subsection{Classifier Inheritance}

We find that the misalignment between the distillation model ${\phi}_d$ and the evaluation model ${\phi}_e$ may lead to some degradation in generalization. Motivated by hypothesis transfer learning \cite{chen2022knowledge, redko2020survey}, which closely matches our scene by leveraging a source-domain classifier for target-domain training with limited labeled data and no access to source data, we inherit the pre-trained classifier $f$ from the distillation model trained on the full images, and focus on learning a projector $ \mathcal{P}$ composed solely of a lightweight linear layer on the synthetic images ${\bm{x}}^{s}$:
\begin{equation}
\begin{aligned}
\mathcal{L}_{eval}=||{\phi}_d(\bm{x}^s)-\mathcal{P}({\phi}_e(\bm{x}^s))||_2^2.
\label{eq:eval}
\end{aligned}
\end{equation}
Notably, label information is no longer required for computing the loss, and the feature alignment loss in Equation \ref{eq:eval} becomes the only source of optimization gradients. We don't need complex fine-tuning of the hyperparameters for different loss weights like soft labels. Once the projector has been trained, we directly reuse the pre-trained classifier for inference on the validation images $\bm{x}^v$:
\begin{equation}
\begin{aligned}
\mathcal{I}=f(\mathcal{P}({\phi}_e(\bm{x}^v))).
\label{eq:infer}
\end{aligned}
\end{equation}
We emphasize that Classifier Inheritance (CI) serves only as an optional plug-in to further improve generalization during evaluation. Unless otherwise specified, all experimental results in this paper are obtained using SFM alone to ensure a fair comparison. More details are provided in the Appendix \ref{app:ci}. 

\section{Experiments}
\subsection{Experimental Setup}
\textbf{Datasets and Backbones.} To evaluate the effectiveness of our method, we conduct all experiments on various datasets with 224 × 224 resolution, including universal ImageNet-1k \cite{deng2009imagenet}, its subsets (ImageFruits, ImageNette, ImageWoof \cite{howard2019smaller}), and fine-grained Stanford Dogs \cite{dataset2011novel} and CUB-200-2011 \cite{welinder2010caltech}. We mainly use the “ViT-B” version of four pre-trained self-supervised models or feature extractors as backbones: CLIP \cite{radford2021learning}, DINO-v2 \cite{oquab2023dinov2}, EVA-02 \cite{fang2024eva}, and MoCo-v3 \cite{chen2021empirical}.

\textbf{Baselines and Evaluation.} We compare our method with the state-of-the-art method LGM \cite{cazenavette2025lgm}, diffusion-based method  MGD$^{3}$ \cite{chan2025mgd}, and three real-image baselines, including \textbf{Random} (simply select a random image for each class), \textbf{Centroids} (take the real image closest to the mean embedding for each class), and \textbf{Neighbors} (choose the real image with the minimum embedding distance to the synthetic image). We use only the final layer of the backbone network to provide transferable features for the task-specific classification head. The reported results represent the averages of 3 trials.



\textbf{Implementation details.} To ensure a fair comparison, we always follow the relevant implementation standards of LGM \cite{cazenavette2025lgm}. Unless otherwise specified, a single differentiable augmentation is applied to our synthetic images. Under the multi-augmentation setting, SFM uses 5 augmentations for CUB-200-2011, 8 for Stanford Dogs, and 10 for ImageNet-100, while LGM uses 3 augmentations for ImageNet-1k and 10 for other datasets. The corresponding LGM* results are cited from the original paper \cite{cazenavette2025lgm}. More implementation details are presented in the Appendix \ref{app:id}.

\input{tables/base}

\subsection{Distillation Performance on Specific Models}
As shown in Table \ref{tab:main}, we distill synthetic images using diverse backbones and evaluate them under consistent architectures. With only one augmentation, our SFM consistently outperforms LGM. For example, on ImageNet-1k with CLIP, SFM achieves 67.1\%, which is 11.6\% higher than LGM (55.5\%) and even 4.1\% higher than LGM* (63.0\%) that uses multiple augmentations. LGM performs extremely poorly or even crashes on fine-grained datasets.  For example, on CUB-200 with EVA-02, SFM reaches 72.5\%, outperforming LGM (51.5\%) by 21\%. Moreover, LGM achieves only 15.0\% and 16.4\% on fine-grained datasets with CLIP, performing even worse than real-image baselines, which further confirms its suboptimality and instability. 

Under multiple augmentations, LGM's inability to capture the global optimal distribution is alleviated to some extent, and the gap with SFM* is significantly reduced, but LGM* still underperforms SFM*. Notably, our single-augmentation Neighbors (SFM) generally outperforms the multi-augmentation Neighbors (LGM*). This indicates that our synthetic images reside in a more representative feature space of the entire dataset, benefiting from our optimal global statistical flow objective.

\input{tables/cross100}
\input{tables/tcdd_ncdd}
\subsection{Generalization Performance Across Models}
As depicted in Table \ref{tab:cross100}, we distill images using a given model and then evaluate them across other models that are not seen during distillation. The generalization of our SFM/SFM*/Neighbors(SFM) generally surpasses that of LGM/LGM*/Neighbors (LGM*). 
We observe that (1) whether used as a distillation model or an evaluation model, the generalization performance of DINO-v2 remains at the highest level, and (2) even when evaluating images distilled using the same model, the performance gap remains significant across different models. For example, when MoCo-v3 is used as the distillation model, DINO-v2 can reach 86.6\%, while CLIP is only 66.3\%, which is far inferior to other real-image baselines. Both cases indicate that the type and quality of self-supervised models can affect the generalization ability of the synthetic dataset. Additionally, we find that CLIP and MoCo‑v3 form an anomalous pair, making them unsuitable for distillation or evaluation, as their performance falls short of the real-image baselines. This is due to the gap and inconsistency in the feature representations learned by the two models \cite{cazenavette2025lgm, huh2024position}.

\subsection{Ablation and Extension}
\textbf{Target Class and Non-target Class}. We distill synthetic images using the two parts in Equation \ref{eq:global_flow} separately, termed Target Class Dataset Distillation (TCDD) and Non-target Class Dataset Distillation (NCDD). As shown in Table \ref{tab:tcdd_ncdd}, NCDD individually collapses across all dataset scales. In contrast, TCDD alone outperforms our SFM (TCDD+NCDD) on the small-scale ImageFruits with only 10 classes, indicating that preserving basic representative features (absolute distribution centers) is sufficient. However, when handling large-scale datasets, our SFM demonstrates greater insight and delivers superior performance gains.  As shown in Figure \ref{fig:tcdd_ncdd}, NCDD deviates entirely from the target-class representation and produces unrelated non-target classes that harm performance, while TCDD effectively captures discriminative features of the target classes and synthesizes images relevant to the target class.  Notably, our synthetic images appear fragmented on the small-scale ImageFruits and are less informative than TCDD, which may explain the performance drop. On the large-scale ImageNet-100, our synthetic images share similar semantics or structures with TCDD, yet our misty quality  brings more expressiveness and insight, making them stand out in evaluation.

\input{tables/extension}
\input{figures/soft_label}
\textbf{Few-shot Learning}. To demonstrate the scalability of our method and its stability under different settings, we extend the extreme 1 IPC to few-shot learning (IPC=2, 5, 10). As shown in Table \ref{tab:ipc}, our SFM can be successfully extended to different IPC and always maintains optimal performance. In contrast, LGM quickly reaches saturation as IPC increases (85.5\% at both IPC=5 and IPC=10), and the gap with Random and MGD${}^{\text{3}}$ narrows (85.5\% vs. 85.0\% and 85.8\%).

\textbf{Classifier Inheritance Helps Model Alignment.} In our previous Table \ref{tab:cross100}, we find that misalignment between the distillation and evaluation models hinders generalization. To address this, we propose an alternative classifier inheritance (CI) strategy that trains the evaluation model directly using the distillation model's extracted features as ``labels”, followed by inference with a classifier pre-trained on the original dataset, ensuring model alignment. As illustrated in Table \ref{tab:ci}, we use DINO-v2 (while MGD${}^\text{3}$ uses the original VAE and diffusion) on ImageNet-100 for distillation and evaluate on other models. Results demonstrate that CI consistently boosts generalization as a plug-in module.

\textbf{Classifier Inheritance Beats Soft Label.} The soft label \cite{qin2024label, sun2024diversity} has been widely established as an effective evaluation method for DD. In Figure \ref{fig:T}, we train our classifier using a weighted loss $\alpha\ell_{kl}+(1-\alpha)\ell_{ce}$, where soft labels are provided by the distillation model along with its corresponding classifier. We observe that soft labels are highly architecture-sensitive. For example, when T=4, they improve performance for DINO-v2 and EVA-02, but they provide no benefit for CLIP and even degrade results for MoCo-v3, as performance drops sharply with increased Kullback–Leibler (KL) divergence loss weight $\alpha$. 
Overall, CI consistently outperforms soft labels, confirming that our inherited classifier is crucial for generalization. Additionally, we find that regardless of how the temperature is adjusted, soft labels consistently fail to surpass our CI. But as T increases, the gap between them becomes smaller. When T=1, soft labels hardly affect performance. In addition, it is worth noting that MoCo-v3 is highly sensitive to both the temperature and the weight $\alpha$ of the soft labels. As T and $\alpha$ increase, the more detrimental the impact of soft labels on performance becomes.

\textbf{Visualization and Explainability}. As shown in Figure \ref{fig:syn_images}, our synthetic images are clearer than those of LGM and exhibit clear structures with no noticeable artifacts, indicating the stability of our global-level SFM during the optimization process. As shown in Figure \ref{fig:flow_both} (a), synthetic images reside on the distribution boundary that carries discriminative information and is distant from the distribution center, which is also why this paradigm is more effective. As shown in Figures \ref{fig:flow_both} (b) and (c), compared to the dynamic process of LGM, our synthetic flow is closer to the statistical flow. This further confirms that our supervised paradigm effectively captures the globally optimal objective.

\input{figures/syn_images}
\input{figures/flow_compare}
\section{Conclusion}
In this work, we are the first to propose the concept of ``flow” in DD, from which we derive a novel paradigm termed Statistical Flow Matching that encourages the synthetic images to efficiently learn the statistical flow of the original dataset through the pre-trained self-supervised model. Based on this, we achieve superior downstream performance using only one image per class, which has a positive impact on edge environments with limitations in computation, storage, and communication. 

\textbf{Limitation and Future Work.} As the number of augmentations increases, our method currently processes all classes simultaneously, which still results in greater GPU memory requirements. Notably, each of our loading steps can be class-independent. Therefore, in future work, we will process classes in blocks of a fixed size, keeping memory usage constant and allowing flexible adjustment of the class block size based on available computational resources.

\begin{ack}
We gratefully acknowledge the financial support from the China Scholarship Council (CSC) (Grant No. 202506070011) for my doctoral studies. This work is supported by the Sichuan Science and Technology Program (Granted No. 2024ZDZX0011, No. 2025ZHCG0002 and No. 2026NSFSC1482).
This work is also supported by Xin Zhang's A*STAR Career Development Fund (CDF) (Project No. H26-KSR0051). 

\end{ack}

\medskip

{\small
\bibliographystyle{abbrv}
\bibliography{neurips_2026}
}

\clearpage

\appendix

\section*{Appendix}
\section{Theoretical Proof}

\subsection{The Proof of Equation \ref{eq:batch_grad}}

The cross-entropy loss $\ell$ is:
\begin{equation}
\bm{\ell} = - \frac{1}{B} \sum_{i=1}^{B} \sum_{c=1}^{C} y_{ic} \log(p_{ic}),
\end{equation}
where $y_{ic}$ is the label value of the $i$-th sample and the $c$-th category.  The predicted probability $p_{ic}$ is:
\begin{equation}
{p}_{ic}=\text{softmax}({z}_{ic})=\frac{\exp\bigl({{z}_{ic}})}{\textstyle\sum_{j=1}^{C}\exp\bigl({{z}_{ij}})},   
\end{equation}
where $\bm{p}_i=[{p}_{i1},{p}_{i2},\cdots ,{p}_{ic},\cdots ,{p}_{iC}]$ is the probability distribution of the $i$-th sample, $\bm{z}_i=[{z}_{i1},{z}_{i2},\cdots ,{z}_{ic},\cdots ,{z}_{iC}]=\bm{W}\bm{\phi}({\bm{x}}_{i})$ is the logits of the $i$-th sample. $\bm{W}$ is the linear classifier matrix.

The gradient of the loss function $\bm{\ell}$ with respect to \( p_{ic} \) is:
\begin{equation}
\frac{\partial \bm{\ell}}{\partial p_{ic}} = - \frac{y_{ic}}{p_{ic}}.
\end{equation}
The gradient of \( p_{ic} \) with respect to \( z_{ic} \) has two cases.

When $c = c'$ (the same class), the derivative is:
\begin{equation}
\begin{aligned}
\frac{\partial {p}_{ic}}{\partial {z}_{ic'}} &=  \partial(\frac{\exp\bigl({z}_{ic})}{\textstyle\sum_{j=1}^{C}\exp\bigl({z}_{ij})})/ \partial {z}_{ic'}^{s} \\ 
&=\frac{\exp\bigl({z}_{ic})}{\textstyle\sum_{j=1}^{C}\exp\bigl({z}_{ij})}       \frac{\textstyle\sum_{j=1}^{C}\exp\bigl({z}_{ic})-\exp\bigl({z}_{ic})}{\textstyle\sum_{j=1}^{C}\exp\bigl({z}_{ij})}  \\
&= {p}_{ic}(1-{p}_{ic}) .
\label{grad3}
\end{aligned}
\end{equation}
 When \( c \neq c' \) (different classes), the derivative is:
\begin{equation}
\begin{aligned}
    \frac{\partial {p}_{ic}}{\partial {z}_{ic'}}  &=  \partial(\frac{\exp\bigl({z}_{ic})} 
 {\textstyle\sum_{j=1}^{C}\exp\bigl({z}_{ij})})/ \partial {z}_{ic'} \\
 &= -\frac{\exp\bigl({z}_{ic'})\exp\bigl({z}_{ic})}{({\textstyle\sum_{j=1}^{C}\exp\bigl({z}_{ij})})^{2}}   \\
 &=-{p}_{ic}\cdot{p}_{ic'} , 
\label{grad2}
\end{aligned}
\end{equation}

For $\bm{\ell}$, the gradient for each part of the logits ${z}_{ic}$ is calculated by chain rule:
\begin{equation}
\begin{aligned}
\frac{\partial \ell}{\partial {z}_{ic}} &= \displaystyle\sum_{j=1}^{C}\frac{\partial  \ell}{\partial {p}_{ij}} \cdot\frac{\partial {p}_{ij}}{\partial {z}_{ic}} \\
&=-\frac{{y}_{ic}}{{p}_{ic}}{p}_{ic}(1-{p}_{ic}) + \displaystyle\sum_{j\ne
i}^{C}{y}_{ij}{p}_{ic}  \\
&=(-{y}_{ic}+{y}_{ic}{p}_{ic}+\displaystyle\sum_{j\ne
i}^{C}{y}_{ij}{p}_{ic}) \\
&=(-{y}_{ic}+{p}_{ic}\displaystyle\sum_{j=
1}^{C}{y}_{ij})  \\
&=({p}_{ic}-{y}_{ic}) \\
\label{grad4}
\end{aligned}
\end{equation}

where ${z}_{ic}=W_c\bm{\phi}({\bm{x}}_{i})$, the gradient of the cross-entropy loss $\bm{\ell}$ with respect to $W_c$ is:
\begin{equation}
\frac{\partial \bm{\ell}}{\partial W_c}= \frac{1}{B} \sum_{i=1}^{B} \frac{\partial \bm{\ell}}{\partial z_{ic}} \cdot \frac{\partial z_{ic}}{\partial W_{c}} = \frac{1}{B} \sum_{i=1}^{B}
(p_{ic} - y_{ic}){\bm{\phi}(x_i)}^{\top},
\end{equation}

Therefore, the gradient of the cross-entropy loss $\bm{\ell}$ with respect to $\bm{W}$ is:
\begin{equation}
\frac{\partial \bm{\ell}}{\partial \bm{W}}= \frac{1}{B} \sum_{i=1}^{B}
(\bm{p}_i - \bm{y}_i){\bm{\phi}(\bm{x}_i})^{\top},
\end{equation}

\subsection{The Proof of Theorem \ref{eq:exchangeable}}

We aim to compute the expectation of the predicted probability \(p_{ic}\) for class \(c\). The softmax function for class \(c\) is defined as:

\begin{equation}
p_{ic} = \frac{e^{W_c^T \bm{\phi}(x_i)}}{\sum_{c'=1}^C e^{W_{c'}^T \bm{\phi}(x_i)}},
\end{equation}

where \(W_c\) is the weight vector for class \(c\) and $\bm{\phi}(x_i)$ is the feature vector for sample \(x_i\). The weight vectors \(W_c\) are independently initialized from a Gaussian distribution $W_c \sim N(0, \sigma_c^2 I)$

We now calculate the expectation of \(p_{ic}\), which is given by:

\begin{equation}
\mathbb{E}[p_{ic}] = \mathbb{E}\left[ \frac{e^{W_c^T \bm{\phi}(x_i)}}{\sum_{c'=1}^C e^{W_{c'}^T \bm{\phi}(x_i)}} \right].
\end{equation}

Since the weight vectors \(W_c\) are independent and identically distributed (i.i.d.) from the same Gaussian distribution, the scoring terms \(W_c^T \bm{\phi}(x_i)\) are symmetric in expectation. Therefore, due to the exchangeability of the weight vectors, the expectation of the class probabilities will be the same across all classes. This implies:

\begin{equation}
\mathbb{E}[e^{W_c^T x_i}] = \mathbb{E}[e^{W_{c'}^T \bm{\phi}(x_i)}] \quad \text{for all} \quad c, c' \in \{1, 2, \dots, C\}.
\end{equation}

Now, considering the symmetry of the weight vectors, the softmax function will normalize the probabilities such that each class will have the same expected probability. Hence, the expected probability for each class is:

\begin{equation}
E[p_{ic}] = \frac{\mathbb{E}[e^{W_c^T \bm{\phi}(x_i)}]}{\mathbb{E}\left[ \sum_{c'=1}^C e^{W_{c'}^T \bm{\phi}(x_i)} \right]} = \frac{1}{C}.
\end{equation}

Therefore, for all classes \(c \in \{1, 2, \dots, C\}\), we have:

\begin{equation}
\mathbb{E}[p_{i1}] = \mathbb{E}[p_{i2}] = \dots = \mathbb{E}[p_{iC}] = \frac{1}{C}.
\end{equation}

\subsection{The Proof of Theorem \ref{th:lognormal}}
\subsection*{1. Derivation of the Expectation of \( X \)}
The expected value of \( X \) is given by $\mathbb{E}[X] = \mathbb{E}[e^Y]$. Since \( Y \sim N(\mu, \sigma^2) \), the probability density function of \( Y \) is:
\begin{equation}
f_Y(y) = \frac{1}{\sqrt{2\pi \sigma^2}} \exp \left( - \frac{(y - \mu)^2}{2\sigma^2} \right),
\end{equation}
Thus, the expectation of \( X \) is:
\begin{equation}
\begin{aligned}
\mathbb{E}[e^Y] &= \int_{-\infty}^{\infty} e^y f_Y(y) \, dy  \\
&= \int_{-\infty}^{\infty} e^y \frac{1}{\sqrt{2\pi \sigma^2}} \exp \left( - \frac{(y - \mu)^2}{2\sigma^2} \right) dy  \\
&= \frac{1}{\sqrt{2\pi \sigma^2}} \int_{-\infty}^{\infty} \exp \left( y - \frac{(y - \mu)^2}{2\sigma^2} \right) dy,
\end{aligned}
\end{equation}

To simplify the integral, we shift the variable by setting \( z = y - \mu \), so \( dz = dy \) and \( y = z + \mu \). This gives:
\begin{equation}
\mathbb{E}[e^Y] = \frac{1}{\sqrt{2\pi \sigma^2}} \int_{-\infty}^{\infty} \exp \left( (z + \mu) - \frac{z^2}{2\sigma^2} \right) dz,
\end{equation}
Now, factor out \( \exp(\mu) \):
\begin{equation}
\mathbb{E}[e^Y] = \frac{1}{\sqrt{2\pi \sigma^2}} \exp(\mu) \int_{-\infty}^{\infty} \exp(z)  \exp \left( - \frac{z^2}{2\sigma^2} \right) dz,
\end{equation}
We know that the Gaussian integral evaluates to:
\begin{equation}
\int_{-\infty}^{\infty}\exp(z) \exp \left( - \frac{z^2}{2\sigma^2} \right) dz = \sqrt{2\pi \sigma^2}\exp ( \frac{\sigma^2}{2}),
\end{equation}
Thus, we obtain the expectation:
\begin{equation}
\mathbb{E}[X] = \mathbb{E}[e^Y] = \exp(\mu) \cdot \exp \left( \frac{\sigma^2}{2} \right) = \exp \left( \mu + \frac{\sigma^2}{2} \right).
\end{equation}

\subsection*{2. Derivation of the Variance of \( X \)}

The variance of \( X \) is given by:
\begin{equation}
\operatorname{Var}[X] = \mathbb{E}[X^2] - (\mathbb{E}[X])^2,
\end{equation}
First, we compute \( \mathbb{E}[X^2] = \mathbb{E}[e^{2Y}] \). The expectation is:
\begin{equation}
\begin{aligned}
\mathbb{E}[e^{2Y}] &= \int_{-\infty}^{\infty} e^{2y} f_Y(y) \, dy \\
&= \int_{-\infty}^{\infty} e^{2y} \frac{1}{\sqrt{2\pi \sigma^2}} \exp \left( - \frac{(y - \mu)^2}{2\sigma^2} \right) dy \\
&= \frac{1}{\sqrt{2\pi \sigma^2}} \int_{-\infty}^{\infty} \exp \left( 2y - \frac{(y - \mu)^2}{2\sigma^2} \right) dy,
\end{aligned}
\end{equation}
 
Shifting the variable again by setting \( z = y - \mu \), we get:
\begin{equation}
\mathbb{E}[e^{2Y}] = \exp(2\mu) \cdot \exp(2\sigma^2),
\end{equation}
Now, we compute the variance:
\begin{equation}
\begin{aligned}
\operatorname{Var}[X] &= \mathbb{E}[e^{2Y}] - \left( \mathbb{E}[e^Y] \right)^2 \\
&= \exp(2\mu) \cdot \exp(2\sigma^2) - \left( \exp(\mu) \cdot \exp \left( \frac{\sigma^2}{2} \right) \right)^2 \\
&= \exp \left( 2\mu + \sigma^2 \right) \left( \exp(\sigma^2) - 1 \right).
\end{aligned}
\end{equation}

\clearpage

\section{More Implementation Details}
\label{app:id}
Following LGM \cite{cazenavette2025lgm}, we train the pyramid representations \cite{fort2025direct} with the Adam optimizer at a learning rate of 0.002. We apply RandomHorizontalFlip (default parameters), RandomResizedCrop (224×224), and RandomGaussianNoise ($\sigma$=0.2) for differentiable augmentation of synthetic images. We distill the synthetic images for 5000 iterations, adding a new level to the pyramid every 200 iterations. During evaluation, the linear classifier is trained for 1000 epochs with a batch size of 100, optimized with Adam. The initial learning rate is set to 0.001/256, which follows the DINO-v1 \cite{caron2021emerging} evaluation protocol, and employs a cosine decay schedule. An early stopping criterion is applied, halting training if test accuracy shows no improvement for 50 epochs. We calculate the statistical center using the total number of images per class as the batch size. If GPU memory is limited, a smaller batch size can be used and yields the same result, demonstrating that our center computation is resource-friendly. We use one H100 GPU for a single augmentation, and two H100 GPUs otherwise.

In this paper, we use an earlier version of LGM, where the Gaussian noise has a mean of 0.5 instead of 0. See \href{https://github.com/GeorgeCazenavette/linear-gradient-matching/commit/33d004b8c16d9ab8e39aec780f9edad767f6f033}{here} for details. This would likely result in darker distilled images since they would have to compensate for the brighter mean noise. In fact, we observe this difference in our experiments. As shown in Table \ref{tab:mean}, changing the mean to 0 indeed leads to a performance improvement.

\input{tables/mean}

\section{Practical Applications or Scenarios}
In our view, dataset distillation, especially in the era of powerful pre-trained self-supervised backbones, matters as a form of data compression and distribution. It transforms part of the learning problem into a resource-aware "how to package and deliver training signals" problem, which becomes valuable when storage, communication, and training costs are bottlenecks. In our setting, which involves very low IPC and frozen pre-trained backbones, concrete scenarios that can benefit include:

\textbf{Edge-device deployment and weak-connectivity settings:} Synthetic datasets are extremely lightweight to store and transmit, enabling faster model updates when bandwidth, storage, or compute is limited.

\textbf{Cross-site sharing and fast reproducibility:} Distilled data can serve as a compact surrogate training set, reducing the burden of dataset management, transfer, and storage for quick verification and reproduction.

\textbf{Fast task bootstrap with strong backbones:} When downstream training is largely “train a lightweight head”, pre-computed distilled sets can shorten iteration cycles by reducing the need to repeatedly curate and distribute large datasets.

\textbf{Privacy-friendly collaboration:} Distilled data does not originate from the original dataset and is not as authentic as the original dataset, thus protecting user privacy to some extent.

\textbf{Extension to richer tasks (future work): } We plan to extend the paradigm to object detection and segmentation, adapting to their task-specific heads, where a small distilled dataset is synthesized on a self-supervised specific backbone and then trained on other downstream backbones.

\section{Classifier Inheritance}
\label{app:ci}
The goal of a neural network in the classification task is to learn a decision boundary that separates different classes in the feature space. In the evaluation of our downstream tasks, as the self-supervised model is frozen, the classifier uniquely learns task-specific information and determines the decision boundary. However, given that our synthetic dataset is highly compressed, such limited data makes it difficult to learn the decision boundary or discriminant information originally acquired by the classifier on the original dataset.  

In our experiments, the projector $ \mathcal{P}$ is extremely lightweight, consisting only of a single linear layer. On the one hand, it aligns the input dimensionality of the golden classifier, and on the other hand, it learns architecture-specific knowledge from the distillation model. Such specific knowledge is crucial for cross-architecture generalization, as the images synthesized according to Equation \ref{eq:sfm} are inherently biased toward the distillation model ${\phi}_d$, and this cannot be eliminated during evaluation. Therefore, we embrace this bias and directly learn the representations of the distillation model. 

\begin{figure*}[t]
    \centering
    \includegraphics[width=\textwidth]{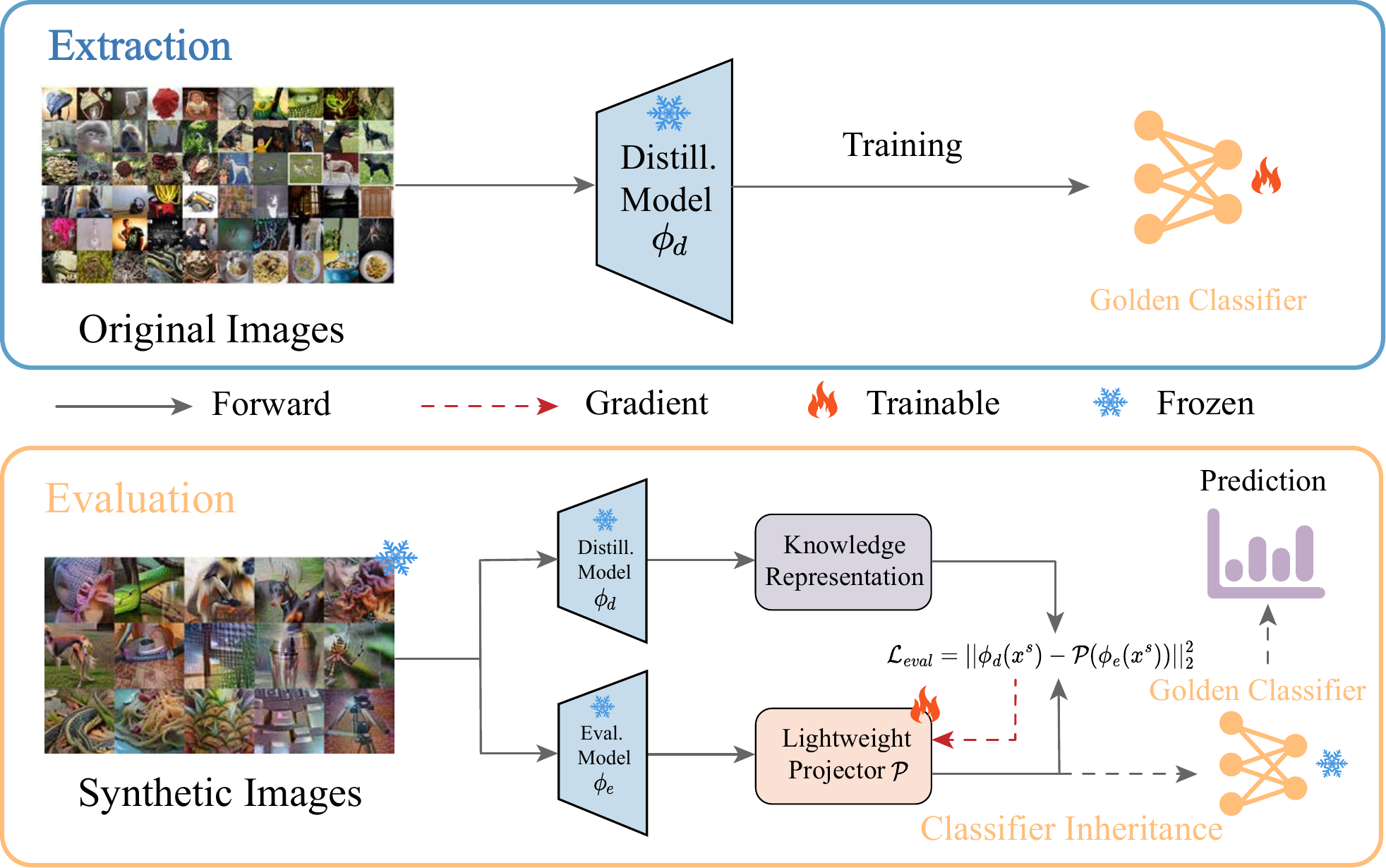}
    \caption{
    \looseness=-1
    An overview of our Classifier Inheritance (CI). We first train a classifier on the original dataset using a frozen distillation model, then freeze it. During evaluation on synthetic images, we directly drive the evaluation model to align its feature representation with that of the distillation model, and use the aforementioned classifier for inference and prediction.
    }
    \label{fig:ci}
\end{figure*}
\input{tables/ci_train}
Our Classifier Inheritance (CI) is simple and storage-efficient, requiring only an additional classifier and one image per class to be stored. This design is well-suited to bandwidth-limited, weakly connected settings and resource-constrained edge environments.

We only train our golden classifier for 10 epochs, and its performance is shown in Table \ref{tab:gc_acc}. Remarkably, it attains high performance with just 10 epochs, highlighting the strong feature extraction ability of pre-trained self-supervised models.

\input{tables/cross1k}
\input{tables/crosscub}
\input{tables/crossdog}

\section{More Experiments}
\textbf{Cross-model Generalization Performance.} We compare the generalization performance of LGM and our SFM on ImageNet-1k, CUB-200-2011, and Stanford Dogs. As shown in Table \ref{tab:cross1k}, Table \ref{tab:crosscub}, and Table \ref{tab:crossdog}, SFM still demonstrates remarkable competitiveness. For example, on ImageNet-1k, when using CLIP as the distillation model, the average generalization performance of SFM is 6\% higher than that of LGM. This also confirms that our method can be seamlessly extended to larger-scale datasets. Distilling ImageNet-1k is memory-intensive. LGM employs 3 rounds of augmentation on synthetic images, which leads to Out-of-Memory (OOM) issues on two H100 GPUs with 90GB of memory each, while our SFM can function normally. However, simultaneously optimizing 1000 synthetic images at each distillation step still imposes high memory demands. We plan to adopt a bi-level optimization paradigm in the future, where the inner loop optimizes a subset of synthetic images and the outer loop iterates through all images, to further reduce memory consumption. This approach may potentially increase training time.

\input{figures/style}
\input{tables/vitsbl}
\textbf{Performance on Homogeneous Architecture.} In previous experiments, we mainly investigated distillation performance on the original model and generalization performance across architectures with large discrepancies (heterogeneous architectures). Beyond that, we also explore generalization across different scales within homogeneous architectures. As shown in Table \ref{tab:vitsbl}, images are distilled using DINO-v2/ViT-B, while the evaluation models are DINO-v2/ViT-S and DINO-v2/ViT-L. As these models are architecturally homogeneous and trained under similar paradigms, they still achieve strong performance with our SFM and CI.

\textbf{Style Transfer.} We distill the art dataset ArtBench \cite{liao2022artbench} and then perform style transfer on top of that. As shown in Figure \ref{fig:style}, we observe that distillation goes beyond simply "copying" samples from the dataset by comparing them to the nearest real neighbor in the model’s embedding space. Instead, distillation involves a more complex transformation, where the model learns to extract and represent the essential features and patterns of the original data. This process leads to a substantial divergence from mere sample replication, as evidenced by the significant differences observed between the distilled data and the original samples in the embedding space. Additionally, we use the synthetic images as a style library for style transfer, transforming the style features of the source image (such as color, texture, brushstrokes, etc.) to generate an image with a new style. We also enhance its artistic expression and visual appeal, creating works that are both realistic and expressive.
\input{tables/gpu}

\textbf{Quantitative Results of Efficiency.} We report the quantitative runtime and GPU consumption under different numbers of Augmentations Per Batch (APB), with EVA-02 as the distillation model. In terms of GPU consumption, SFM requires about 0.5 GB less memory than LGM. This mainly stems from the fact that LGM keeps the original-data dataloader resident in memory and retrieves samples from it at every distillation step, whereas SFM relies only on globally fixed statistics. In terms of runtime, LGM is substantially slower than SGM due to frequent data fetching and heavier gradient computations, and this gap further widens as APB increases.


\newpage

\end{document}

%% file: figures/motivation.tex
\begin{wrapfigure}{r}{0.5\textwidth}
    \vspace{-6pt}
    \centering
    \includegraphics[width=1.00\linewidth]{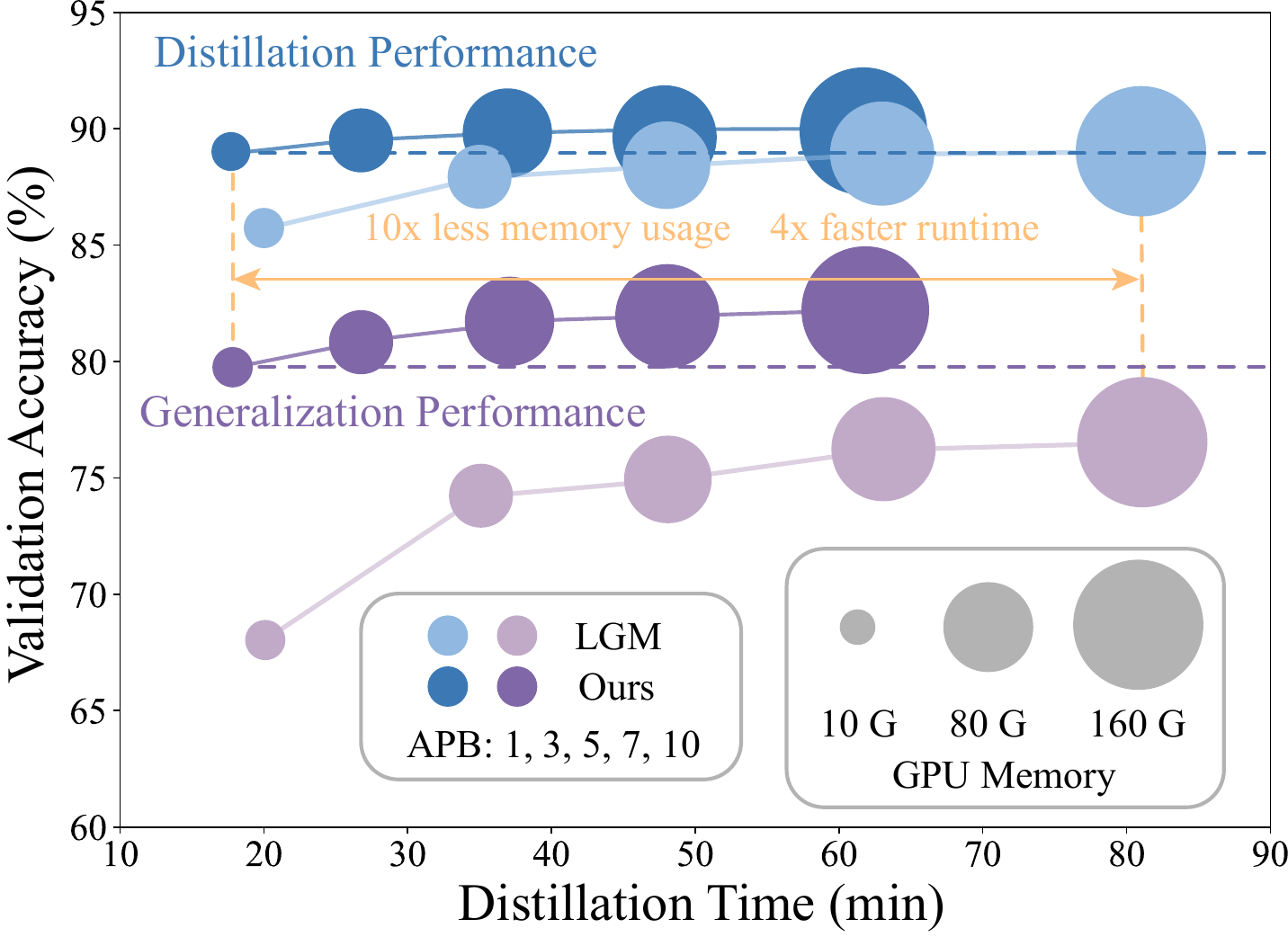}
    \caption{
    Comparison between LGM and our SFM in terms of distillation time, GPU memory usage, and validation accuracy on ImageNet-100 with IPC=1. The size of the circle denotes different amounts of Augmentations Per Batch (APB). The distillation model is EVA-02 and the generalized models are CLIP, DINO-v2 and MoCo-v3. With only one augmentation, our SFM beats LGM using multiple augmentations, which substantially reduces resource consumption while achieving state-of-the-art validation performance.}
    \label{fig:motivation}
    \vspace{-10pt}
\end{wrapfigure}

%% file: figures/overview.tex
\begin{figure*}[t]
    \centering
    \includegraphics[width=\textwidth]{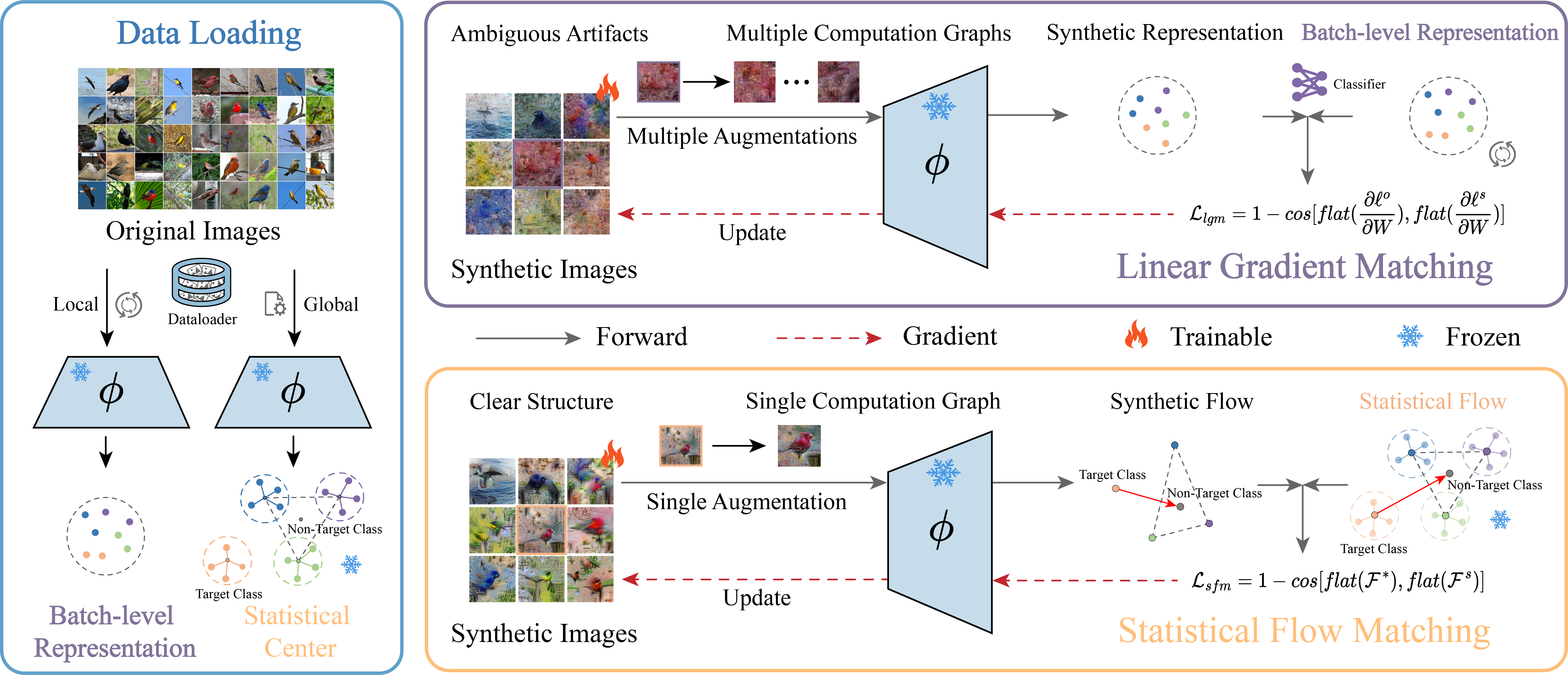}
    \caption{
    \looseness=-1
    Comparison of the pipelines: Linear Gradient Matching vs. Statistical Flow Matching. During the original data loading phase, LGM dynamically loads a large batch of samples at every distillation step per epoch, whereas SFM computes the statistics only once and then fixes them during subsequent optimization. During the synthetic data distillation phase, LGM employs multiple augmentations to effectively align gradients or capture local relative distributions, while retaining expensive model computation graphs for backpropagation. In contrast, SFM requires only a single augmentation to stably and efficiently match the optimal statistical flow of the original data.
    }
    \label{fig:overview}
\end{figure*}

%% file: tables/analytic.tex
\begin{wraptable}{r}{0.5\textwidth}
\vspace{-10pt}
\small
\centering
\setlength{\tabcolsep}{3.5pt}
\renewcommand{\arraystretch}{1.3}
\caption{A comparison of average generalization performance under different \( \bm{W}\) settings. The number of augmentation per batch is set to 1. It proves to be insensitive to changes in \( \bm{W}\). }
\label{tab:analytic}
    \begin{tabular}{lcccc}
    \hline
    
    Mode     & CLIP                              & DINO-v2                             & EVA-02                               & MoCo-v3   \\  \hline
    Random            & 64.9\smaller{\gray{$\pm$0.2}}     & 78.9\smaller{\gray{$\pm$0.1}}       & 72.2\smaller{\gray{$\pm$0.1}}        & 72.5\smaller{\gray{$\pm$0.1}}             \\
    Fixed             & 64.4\smaller{\gray{$\pm$0.4}}     & 78.4\smaller{\gray{$\pm$0.2}}       & 72.6\smaller{\gray{$\pm$0.3}}        & 72.4\smaller{\gray{$\pm$0.2}}         \\
    Analytic          & 64.3\smaller{\gray{$\pm$0.4}}     & 78.4\smaller{\gray{$\pm$0.2}}       & 73.0\smaller{\gray{$\pm$0.1}}        & 71.9\smaller{\gray{$\pm$0.2}}        \\ \hline
\end{tabular}
\end{wraptable}

%% file: tables/base.tex
\begin{table}
\smaller
\setlength{\tabcolsep}{0pt}
	\caption{\textbf{The distillation performance on specific models.} We compare our SFM with LGM and several real-image baselines on various datasets. Our SFM, along with its derivatives SFM* and Neighbors (SFM), consistently outperforms other relevant baselines. The superscript ``*” indicates multiple augmentations, otherwise a single augmentation. ``*'' indicates out-of-memory due to resource limitations. The best and second-best results are highlighted in \textbf{bold} and \underline{underlined}.}
	\begin{center}{
		\begin{tabularx}{1.0\textwidth}{lYYYYYYYYYY}
			\toprule\multirow{2}{*}{\shortstack{Train Set\\(1 Img/Cls)}} & \multicolumn{ 5}{c}{\normalsize ImageNet-100} & \multicolumn{ 5}{c}{\normalsize ImageNet-1k}\\\cmidrule(lr){2-6}\cmidrule(lr){7-11}
			& {CLIP} & {DINO-v2} & {EVA-02} & {MoCo-v3} & {Average} & {CLIP} & {DINO-v2} & {EVA-02} & {MoCo-v3} & {Average} \\ \midrule
            Random	& 56.6\smaller{\gray{$\pm$1.6}}	& 74.8\smaller{\gray{$\pm$2.8}}	& 64.5\smaller{\gray{$\pm$2.7}}	& 61.4\smaller{\gray{$\pm$2.6}}	& 64.3\smaller{\gray{$\pm$2.4}}	& 31.7\smaller{\gray{$\pm$0.5}}	& 50.3\smaller{\gray{$\pm$0.5}}	& 37.7\smaller{\gray{$\pm$0.4}}	& 38.8\smaller{\gray{$\pm$0.6}}	& 39.6\smaller{\gray{$\pm$0.5}}	\\
			Centroids	& 77.1\smaller{\gray{$\pm$0.1}}	& 86.9\smaller{\gray{$\pm$0.3}}	& 80.9\smaller{\gray{$\pm$0.2}}	& 77.7\smaller{\gray{$\pm$0.1}}	& 80.6\smaller{\gray{$\pm$0.2}}	& 53.9\smaller{\gray{$\pm$0.0}}	& 69.5\smaller{\gray{$\pm$0.1}}	& 58.1\smaller{\gray{$\pm$0.1}}	& 57.4\smaller{\gray{$\pm$0.0}}	& 59.7\smaller{\gray{$\pm$0.1}}	\\
            Neighbors (LGM*)	& 67.8\smaller{\gray{$\pm$0.3}}	& 86.0\smaller{\gray{$\pm$0.2}}	& 78.8\smaller{\gray{$\pm$0.2}}	& 77.1\smaller{\gray{$\pm$0.1}}	& 77.4\smaller{\gray{$\pm$0.2}}	& 38.8\smaller{\gray{$\pm$0.1}}	& 67.7\smaller{\gray{$\pm$0.1}}	& 49.9\smaller{\gray{$\pm$0.1}}	& 56.4\smaller{\gray{$\pm$0.0}}	& 53.2\smaller{\gray{$\pm$0.1}}	\\
            \textbf{Neighbors (SFM)}	& 72.2\smaller{\gray{$\pm$0.2}}	& 88.0\smaller{\gray{$\pm$0.1}}	& 80.4\smaller{\gray{$\pm$0.0}}	& 77.8\smaller{\gray{$\pm$0.1}}	& 79.6\smaller{\gray{$\pm$0.1}}	& 48.1\smaller{\gray{$\pm$0.1}}	& 69.6\smaller{\gray{$\pm$0.0}}	& 57.3\smaller{\gray{$\pm$0.0}}	& 57.0\smaller{\gray{$\pm$0.0}}	& 58.0\smaller{\gray{$\pm$0.0}}	\\
            LGM	& 77.3\smaller{\gray{$\pm$0.2}}	& 89.8\smaller{\gray{$\pm$0.0}}	& 85.7\smaller{\gray{$\pm$0.1}}	& 81.1\smaller{\gray{$\pm$0.1}}	& 83.5\smaller{\gray{$\pm$0.1}}	& 55.5\smaller{\gray{$\pm$0.0}}	& 73.5\smaller{\gray{$\pm$0.2}}	& 67.6\smaller{\gray{$\pm$0.0}}	& 61.6\smaller{\gray{$\pm$0.0}}	& 64.6\smaller{\gray{$\pm$0.1}}	\\ 
            \textbf{SFM (Ours)}	& \underline{86.7}\smaller{\gray{$\pm$0.1}}	& 90.6\smaller{\gray{$\pm$0.0}}	& 88.9\smaller{\gray{$\pm$0.2}}	& \underline{84.7}\smaller{\gray{$\pm$0.0}}	& \underline{87.7}\smaller{\gray{$\pm$0.1}}	& \textbf{67.1}\smaller{\gray{$\pm$0.0}}	& \textbf{75.1}\smaller{\gray{$\pm$0.1}}	& \textbf{71.6}\smaller{\gray{$\pm$0.1}}	& \textbf{63.4}\smaller{\gray{$\pm$0.0}}	& \textbf{69.3}\smaller{\gray{$\pm$0.1}}	\\ 
            LGM*	& 84.9\smaller{\gray{$\pm$0.1}}	& \underline{91.5}\smaller{\gray{$\pm$0.1}}	& \underline{89.0}\smaller{\gray{$\pm$0.0}}	& 83.4\smaller{\gray{$\pm$0.1}}	& 87.2\smaller{\gray{$\pm$0.1}}	& \underline{63.0}\smaller{\gray{$\pm$0.0}}	& \underline{75.0}\smaller{\gray{$\pm$0.1}}	& \underline{70.3}\smaller{\gray{$\pm$0.1}}	& \underline{63.2}\smaller{\gray{$\pm$0.0}}	& \underline{67.9}\smaller{\gray{$\pm$0.0}}	\\
            \textbf{SFM* (Ours)}	& \textbf{86.9\smaller{\gray{$\pm$0.1}}}	& \textbf{91.6\smaller{\gray{$\pm$0.1}}}	& \textbf{89.7\smaller{\gray{$\pm$0.2}}}	& \textbf{85.6\smaller{\gray{$\pm$0.0}}}	& \textbf{88.5\smaller{\gray{$\pm$0.1}}}	& --	& --	& --	& --	& --	\\ \midrule 
			Full Dataset	& 92.5\smaller{\gray{$\pm$0.0}}	& 95.2\smaller{\gray{$\pm$0.1}}	& 94.1\smaller{\gray{$\pm$0.1}}	& 89.4\smaller{\gray{$\pm$0.3}}	& 92.8\smaller{\gray{$\pm$0.1}}	& 78.7\smaller{\gray{$\pm$0.0}}	& 83.0\smaller{\gray{$\pm$0.0}}	& 81.7\smaller{\gray{$\pm$0.1}}	& 76.5\smaller{\gray{$\pm$0.0}}	& 80.0\smaller{\gray{$\pm$0.0}}	\\ \midrule
           \multirow{2}{*}{\shortstack{Train Set\\(1 Img/Cls)}} & \multicolumn{5}{c}{\normalsize Stanford Dogs} & \multicolumn{5}{c}{\normalsize CUB-200}\\\cmidrule(lr){2-6}\cmidrule(lr){7-11}
			& {CLIP} & {DINO-v2} & {EVA-02} & {MoCo-v3} & {Average} & {CLIP} & {DINO-v2} & {EVA-02} & {MoCo-v3} & {Average} \\ \midrule
            Random	& 23.3\smaller{\gray{$\pm$1.5}}	& 51.9\smaller{\gray{$\pm$1.8}}	& 38.3\smaller{\gray{$\pm$1.8}}	& 36.6\smaller{\gray{$\pm$1.4}}	& 37.5\smaller{\gray{$\pm$1.6}}	& 37.5\smaller{\gray{$\pm$1.6}}	& 64.4\smaller{\gray{$\pm$1.5}}	& 44.3\smaller{\gray{$\pm$1.5}}	& 19.1\smaller{\gray{$\pm$0.5}}	& 41.3\smaller{\gray{$\pm$1.3}}	\\
            Centroids	& 43.3\smaller{\gray{$\pm$0.1}}	& 73.0\smaller{\gray{$\pm$0.2}}	& 60.9\smaller{\gray{$\pm$0.2}}	& 55.2\smaller{\gray{$\pm$0.2}}	& 58.1\smaller{\gray{$\pm$0.2}}	& 54.3\smaller{\gray{$\pm$0.2}}	& 78.5\smaller{\gray{$\pm$0.2}}	& 59.9\smaller{\gray{$\pm$0.3}}	& 30.2\smaller{\gray{$\pm$0.1}}	& 55.7\smaller{\gray{$\pm$0.2}}	\\
            Neighbors (LGM*)	& 33.4\smaller{\gray{$\pm$0.1}}	& 71.3\smaller{\gray{$\pm$0.2}}	& 58.5\smaller{\gray{$\pm$0.2}}	& 56.3\smaller{\gray{$\pm$0.1}}	& 54.9\smaller{\gray{$\pm$0.2}}	& 39.4\smaller{\gray{$\pm$0.1}}	& 76.9\smaller{\gray{$\pm$0.0}}	& 52.6\smaller{\gray{$\pm$0.3}}	& 28.1\smaller{\gray{$\pm$0.0}}	& 49.2\smaller{\gray{$\pm$0.1}}	\\
            \textbf{Neighbors (SFM)}	& 35.7\smaller{\gray{$\pm$0.1}}	& 72.3\smaller{\gray{$\pm$0.2}}	& 60.3\smaller{\gray{$\pm$0.2}}	& 55.5\smaller{\gray{$\pm$0.3}}	& 56.0\smaller{\gray{$\pm$0.2}}	& 47.3\smaller{\gray{$\pm$0.3}}	& 78.7\smaller{\gray{$\pm$0.1}}	& 56.0\smaller{\gray{$\pm$0.2}}	& 28.7\smaller{\gray{$\pm$0.1}}	& 52.7\smaller{\gray{$\pm$0.2}}	\\
			LGM	& 15.0\smaller{\gray{$\pm$0.0}}	& 79.8\smaller{\gray{$\pm$0.2}}	& 65.0\smaller{\gray{$\pm$0.2}}	& 63.5\smaller{\gray{$\pm$0.2}}	& 55.8\smaller{\gray{$\pm$0.2}}	& 16.4\smaller{\gray{$\pm$0.0}}	& 81.1\smaller{\gray{$\pm$0.1}}	& 51.5\smaller{\gray{$\pm$0.2}}	& 25.5\smaller{\gray{$\pm$0.0}}	& 43.6\smaller{\gray{$\pm$0.1}}	\\
            \textbf{SFM (Ours)}	& \underline{55.0}\smaller{\gray{$\pm$0.3}}	& 81.8\smaller{\gray{$\pm$0.2}}	& 72.9\smaller{\gray{$\pm$0.2}}	& 67.2\smaller{\gray{$\pm$0.3}}	& 69.2\smaller{\gray{$\pm$0.3}}	& \underline{67.6}\smaller{\gray{$\pm$0.2}}	& 85.7\smaller{\gray{$\pm$0.1}}	& 72.5\smaller{\gray{$\pm$0.2}}	& 41.4\smaller{\gray{$\pm$0.1}}	& \underline{66.8}\smaller{\gray{$\pm$0.2}}	\\
            LGM*	& 52.1\smaller{\gray{$\pm$0.2}}	& \underline{83.0}\smaller{\gray{$\pm$0.1}}	& \underline{74.8}\smaller{\gray{$\pm$0.1}}	& \underline{69.6}\smaller{\gray{$\pm$0.2}}	& \underline{69.9}\smaller{\gray{$\pm$0.2}}	& 62.2\smaller{\gray{$\pm$0.2}}	& \underline{86.0}\smaller{\gray{$\pm$0.1}}	& \underline{74.1}\smaller{\gray{$\pm$0.2}}	& \underline{42.5\smaller{\gray{$\pm$0.2}}}	& 66.2\smaller{\gray{$\pm$0.2}}	\\
           \textbf{SFM* (Ours)}	& \textbf{58.9\smaller{\gray{$\pm$0.0}}}	& \textbf{83.2\smaller{\gray{$\pm$0.1}}}	& \textbf{75.0\smaller{\gray{$\pm$0.1}}}	& \textbf{70.4\smaller{\gray{$\pm$0.2}}}	& \textbf{71.9\smaller{\gray{$\pm$0.1}}}	& \textbf{70.5\smaller{\gray{$\pm$0.1}}}	& \textbf{86.5\smaller{\gray{$\pm$0.0}}}	& \textbf{75.0\smaller{\gray{$\pm$0.0}}}	& \textbf{45.5\smaller{\gray{$\pm$0.1}}}	& \textbf{69.4\smaller{\gray{$\pm$0.1}}}	\\ \midrule 
			Full Dataset	& 76.9\smaller{\gray{$\pm$0.1}}	& 88.6\smaller{\gray{$\pm$0.1}}	& 82.6\smaller{\gray{$\pm$0.1}}	& 72.3\smaller{\gray{$\pm$0.5}}	& 80.1\smaller{\gray{$\pm$0.2}}	& 77.5\smaller{\gray{$\pm$0.7}}	& 90.2\smaller{\gray{$\pm$0.2}}	& 84.0\smaller{\gray{$\pm$0.3}}	& 43.7\smaller{\gray{$\pm$0.8}}	& 73.8\smaller{\gray{$\pm$0.5}}	\\\bottomrule
		\end{tabularx}
	}\end{center} 
	\label{tab:main}
\end{table}

%% file: tables/cross100.tex
\begin{table}\smaller\setlength{\tabcolsep}{0pt}
	\caption{\textbf{The generalization performance across models.} We synthesize images using a given model and then evaluate them across other models on ImageNet-100. Compared to other baselines, our distilled datasets generalize well and achieve optimal performance across different model pairs, aside from outlier pairs of CLIP and MoCo-v3, which is caused by discrepancies or inconsistencies inherent in the models themselves. The superscript ``*'' indicates multiple augmentations, otherwise a single augmentation. The best and second-best results are highlighted in \textbf{bold} and \underline{underlined}.}
	\begin{center}{
		\begin{tabularx}{1.0\textwidth}{lYYYYYYYYYY}
        \toprule
			\shortstack{Distill. Model} & \multicolumn{ 5}{c}{\normalsize CLIP} & \multicolumn{ 5}{c}{\normalsize DINO-v2}\\\cmidrule(lr){2-6}\cmidrule(lr){7-11}
			\shortstack{Eval. Model}& {CLIP} & {DINO-v2} & {EVA-02} & {MoCo-v3} & {Average} & {CLIP} & {DINO-v2} & {EVA-02} & {MoCo-v3} & {Average} \\ \midrule
			Centroids	& \gray{77.1}\smaller{\gray{$\pm$0.1}}	& \underline{85.1}\smaller{\gray{$\pm$0.2}}	& 82.6\smaller{\gray{$\pm$0.2}}	& \textbf{74.7}\smaller{\gray{$\pm$0.1}}	& 79.9\smaller{\gray{$\pm$0.2}}	& 72.4\smaller{\gray{$\pm$0.0}}	& \gray{86.9}\smaller{\gray{$\pm$0.3}}	& 80.1\smaller{\gray{$\pm$0.4}}	& 75.8\smaller{\gray{$\pm$0.1}}	& 78.8\smaller{\gray{$\pm$0.2}}	\\ 
            Neighbors (LGM*)	& \gray{67.8}\smaller{\gray{$\pm$0.3}}	& 81.1\smaller{\gray{$\pm$0.1}}	& 75.0\smaller{\gray{$\pm$0.2}}	& 70.4\smaller{\gray{$\pm$0.0}}	& 73.5\smaller{\gray{$\pm$0.1}}	& 71.2\smaller{\gray{$\pm$0.1}}	& \gray{86.0}\smaller{\gray{$\pm$0.2}}	& 78.0\smaller{\gray{$\pm$0.4}}	& 74.7\smaller{\gray{$\pm$0.0}}	& 77.6\smaller{\gray{$\pm$0.2}}	\\
           \textbf{Neighbors (SFM)}	& \gray{72.2}\smaller{\gray{$\pm$0.2}}	& 83.6\smaller{\gray{$\pm$0.2}}	& 75.8\smaller{\gray{$\pm$0.0}}	& \underline{72.3}\smaller{\gray{$\pm$0.1}}	& 76.0\smaller{\gray{$\pm$0.1}}	& 72.7\smaller{\gray{$\pm$0.0}}	& \gray{88.0}\smaller{\gray{$\pm$0.1}}	& 80.0\smaller{\gray{$\pm$0.2}}	& 75.4\smaller{\gray{$\pm$0.0}}	& 79.0\smaller{\gray{$\pm$0.1}}	\\
            LGM	& \gray{77.3}\smaller{\gray{$\pm$0.2}}	& 70.2\smaller{\gray{$\pm$0.4}}	& 70.4\smaller{\gray{$\pm$0.1}}	& 41.8\smaller{\gray{$\pm$0.2}}	& 64.9\smaller{\gray{$\pm$0.2}}	& 70.8\smaller{\gray{$\pm$0.1}}	& \gray{89.8}\smaller{\gray{$\pm$0.0}}	& 82.4\smaller{\gray{$\pm$0.2}}	& 72.5\smaller{\gray{$\pm$0.1}}	& 78.9\smaller{\gray{$\pm$0.1}}	\\ 
            \textbf{SFM (Ours)}	& \underline{\gray{86.7}}\smaller{\gray{$\pm$0.1}}	& 84.9\smaller{\gray{$\pm$0.3}}	& \underline{85.5}\smaller{\gray{$\pm$0.2}}	& 68.7\smaller{\gray{$\pm$0.0}}	& \underline{81.5}\smaller{\gray{$\pm$0.2}}	& \underline{78.6}\smaller{\gray{$\pm$0.0}}	& \gray{90.6}\smaller{\gray{$\pm$0.0}}	& 86.2\smaller{\gray{$\pm$0.1}}	& \underline{80.5}\smaller{\gray{$\pm$0.0}}	& \underline{84.0}\smaller{\gray{$\pm$0.0}}	\\
            LGM*	& \gray{84.9}\smaller{\gray{$\pm$0.1}}	& 80.8\smaller{\gray{$\pm$0.4}}	& 83.8\smaller{\gray{$\pm$0.2}}	& 61.6\smaller{\gray{$\pm$0.2}}	& 77.8\smaller{\gray{$\pm$0.2}}	& 77.0\smaller{\gray{$\pm$0.1}}	& \underline{\gray{91.5}}\smaller{\gray{$\pm$0.1}}	& \underline{86.8}\smaller{\gray{$\pm$0.1}}	& 78.8\smaller{\gray{$\pm$0.1}}	& 83.5\smaller{\gray{$\pm$0.1}}	\\
            \textbf{SFM* (Ours)}	& \textbf{\gray{86.9}\smaller{\gray{$\pm$0.1}}}	& \textbf{86.5\smaller{\gray{$\pm$0.2}}}	& \textbf{87.5\smaller{\gray{$\pm$0.1}}}	& 72.0\smaller{\gray{$\pm$0.1}}	& \textbf{83.2\smaller{\gray{$\pm$0.1}}}	& \textbf{79.8\smaller{\gray{$\pm$0.0}}}	& \textbf{\gray{91.6}\smaller{\gray{$\pm$0.1}}}	& \textbf{87.1\smaller{\gray{$\pm$0.1}}}	& \textbf{81.4\smaller{\gray{$\pm$0.0}}}	& \textbf{85.0\smaller{\gray{$\pm$0.1}}}	\\ \midrule
			\shortstack{Distill. Model} & \multicolumn{ 5}{c}{\normalsize EVA-02} & \multicolumn{ 5}{c}{\normalsize MoCo-v3}\\\cmidrule(lr){2-6}\cmidrule(lr){7-11}
			\shortstack{Eval. Model}& {CLIP} & {DINO-v2} & {EVA-02} & {MoCo-v3} & {Average} & {CLIP} & {DINO-v2} & {EVA-02} & {MoCo-v3} & {Average} \\ \midrule
			Centroids	& 70.7\smaller{\gray{$\pm$0.1}}	& 85.2\smaller{\gray{$\pm$0.1}}	& \gray{80.9}\smaller{\gray{$\pm$0.2}}	& 72.6\smaller{\gray{$\pm$0.1}}	& 77.4\smaller{\gray{$\pm$0.1}}	& \underline{75.5}\smaller{\gray{$\pm$0.2}}	& 85.8\smaller{\gray{$\pm$0.1}}	& 81.7\smaller{\gray{$\pm$0.1}}	& \gray{77.7}\smaller{\gray{$\pm$0.1}}	& \underline{80.2}\smaller{\gray{$\pm$0.1}}	\\
            Neighbors (LGM*)	& 69.5\smaller{\gray{$\pm$0.1}}	& 85.4\smaller{\gray{$\pm$0.3}}	& \gray{78.8}\smaller{\gray{$\pm$0.2}}	& 72.3\smaller{\gray{$\pm$0.1}}	& 76.5\smaller{\gray{$\pm$0.2}}	& 72.6\smaller{\gray{$\pm$0.3}}	& 86.2\smaller{\gray{$\pm$0.2}}	& 81.1\smaller{\gray{$\pm$0.5}}	& \gray{77.1}\smaller{\gray{$\pm$0.1}}	& 79.7\smaller{\gray{$\pm$0.3}}	\\
           \textbf{Neighbors (SFM)}& 69.4\smaller{\gray{$\pm$0.3}}	& 85.2\smaller{\gray{$\pm$0.1}}	& \gray{80.4}\smaller{\gray{$\pm$0.0}}	& \underline{74.7}\smaller{\gray{$\pm$0.1}}	& 77.4\smaller{\gray{$\pm$0.1}}	& \textbf{75.8}\smaller{\gray{$\pm$0.0}}	& 85.7\smaller{\gray{$\pm$0.1}}	& 81.2\smaller{\gray{$\pm$0.1}}	& \gray{77.8}\smaller{\gray{$\pm$0.1}}	& 80.1\smaller{\gray{$\pm$0.1}}	\\
            LGM	& 69.7\smaller{\gray{$\pm$0.0}}	& 81.4\smaller{\gray{$\pm$0.3}}	& \gray{85.7}\smaller{\gray{$\pm$0.1}}	& 52.0\smaller{\gray{$\pm$0.0}}	& 72.2\smaller{\gray{$\pm$0.1}}	& 54.3\smaller{\gray{$\pm$0.1}}	& 81.7\smaller{\gray{$\pm$0.1}}	& 72.9\smaller{\gray{$\pm$0.1}}	& \gray{81.1}\smaller{\gray{$\pm$0.1}}	& 72.5\smaller{\gray{$\pm$0.1}}	\\ 
            \textbf{SFM (Ours)}	& \underline{81.0}\smaller{\gray{$\pm$0.2}}	& \underline{88.0}\smaller{\gray{$\pm$0.2}}	& \gray{88.9}\smaller{\gray{$\pm$0.2}}	& 70.5\smaller{\gray{$\pm$0.0}}	& \underline{82.1}\smaller{\gray{$\pm$0.2}}	& 66.3\smaller{\gray{$\pm$0.1}}	& \underline{86.7}\smaller{\gray{$\pm$0.2}}	& 79.3\smaller{\gray{$\pm$0.2}}	& \underline{\gray{84.7}}\smaller{\gray{$\pm$0.0}}	& 79.3\smaller{\gray{$\pm$0.1}}	\\
            LGM*	& 75.5\smaller{\gray{$\pm$0.2}}	& 86.4\smaller{\gray{$\pm$0.1}}	& \underline{\gray{89.0}}\smaller{\gray{$\pm$0.0}}	& 67.7\smaller{\gray{$\pm$0.1}}	& 79.7\smaller{\gray{$\pm$0.1}}	& 65.6\smaller{\gray{$\pm$0.1}}	& 86.6\smaller{\gray{$\pm$0.1}}	& \underline{82.3}\smaller{\gray{$\pm$0.2}}	& \gray{83.4}\smaller{\gray{$\pm$0.1}}	& 79.5\smaller{\gray{$\pm$0.1}}	\\
             \textbf{SFM* (Ours)}	& \textbf{81.5\smaller{\gray{$\pm$0.1}}}	& \textbf{88.9\smaller{\gray{$\pm$0.2}}}	& \textbf{\gray{89.7}\smaller{\gray{$\pm$0.2}}}	& \textbf{75.5\smaller{\gray{$\pm$0.0}}}	& \textbf{83.9\smaller{\gray{$\pm$0.1}}}	& 72.5\smaller{\gray{$\pm$0.0}}	& \textbf{88.9\smaller{\gray{$\pm$0.2}}}	& \textbf{84.5\smaller{\gray{$\pm$0.2}}}	& \textbf{\gray{85.6}\smaller{\gray{$\pm$0.0}}}	& \textbf{82.9\smaller{\gray{$\pm$0.1}}}	\\ \midrule 
			Full Dataset	& 92.5\smaller{\gray{$\pm$0.0}}	& 95.2\smaller{\gray{$\pm$0.1}}	& 94.1\smaller{\gray{$\pm$0.1}}	& 89.4\smaller{\gray{$\pm$0.3}}	& 92.8\smaller{\gray{$\pm$0.1}}	& 92.5\smaller{\gray{$\pm$0.0}}	& 95.2\smaller{\gray{$\pm$0.1}}	& 94.1\smaller{\gray{$\pm$0.1}}	& 89.4\smaller{\gray{$\pm$0.3}}	& 92.8\smaller{\gray{$\pm$0.1}}	\\ \bottomrule
		\end{tabularx}
	}\end{center} 
    \vspace{-10pt}
	\label{tab:cross100}
\end{table}

%% file: tables/tcdd_ncdd.tex
\begin{figure}[t]
    \centering
    \begin{minipage}[b]{0.45\textwidth}\setlength{\tabcolsep}{5pt}
        \centering\smaller             
        \renewcommand{\arraystretch}{1.2}
        \captionof{table}{The Independent and combined effects of Target Class (TCDD) and Non-target Class (NCDD) Dataset Distillation on datasets of varying scales. The results represent the average cross-architecture generalization performance distilled from CLIP.}
        \begin{tabularx}{\textwidth}{ccccc}
        \hline
        TCDD              & NCDD                      & IN-Fruits                                & IN-100                                    & IN-1k                                              \\  \hline
        $\times$          & $\checkmark$              & 15.8\smaller{\gray{$\pm$1.2}}             & 0.77\smaller{\gray{$\pm$0.0}}              & 0.10\smaller{\gray{$\pm$0.1}}                 \\
        $\checkmark$      & $\times$                  & \textbf{77.6\smaller{\gray{$\pm$0.7}}}           & 79.4\smaller{\gray{$\pm$0.1}}             & 57.9\smaller{\gray{$\pm$0.0}}                \\
        $\checkmark$      & $\checkmark$              & 76.5\smaller{\gray{$\pm$1.1}}   & \textbf{81.5\smaller{\gray{$\pm$0.2}}}        & \textbf{60.7\smaller{\gray{$\pm$0.1}}}                      \\ \hline
        \end{tabularx}
        \vspace{-1em}
        \label{tab:tcdd_ncdd}
    \end{minipage}
    \hfill
    \begin{minipage}[b]{0.5\textwidth}
        \centering
        \includegraphics[width=\textwidth]{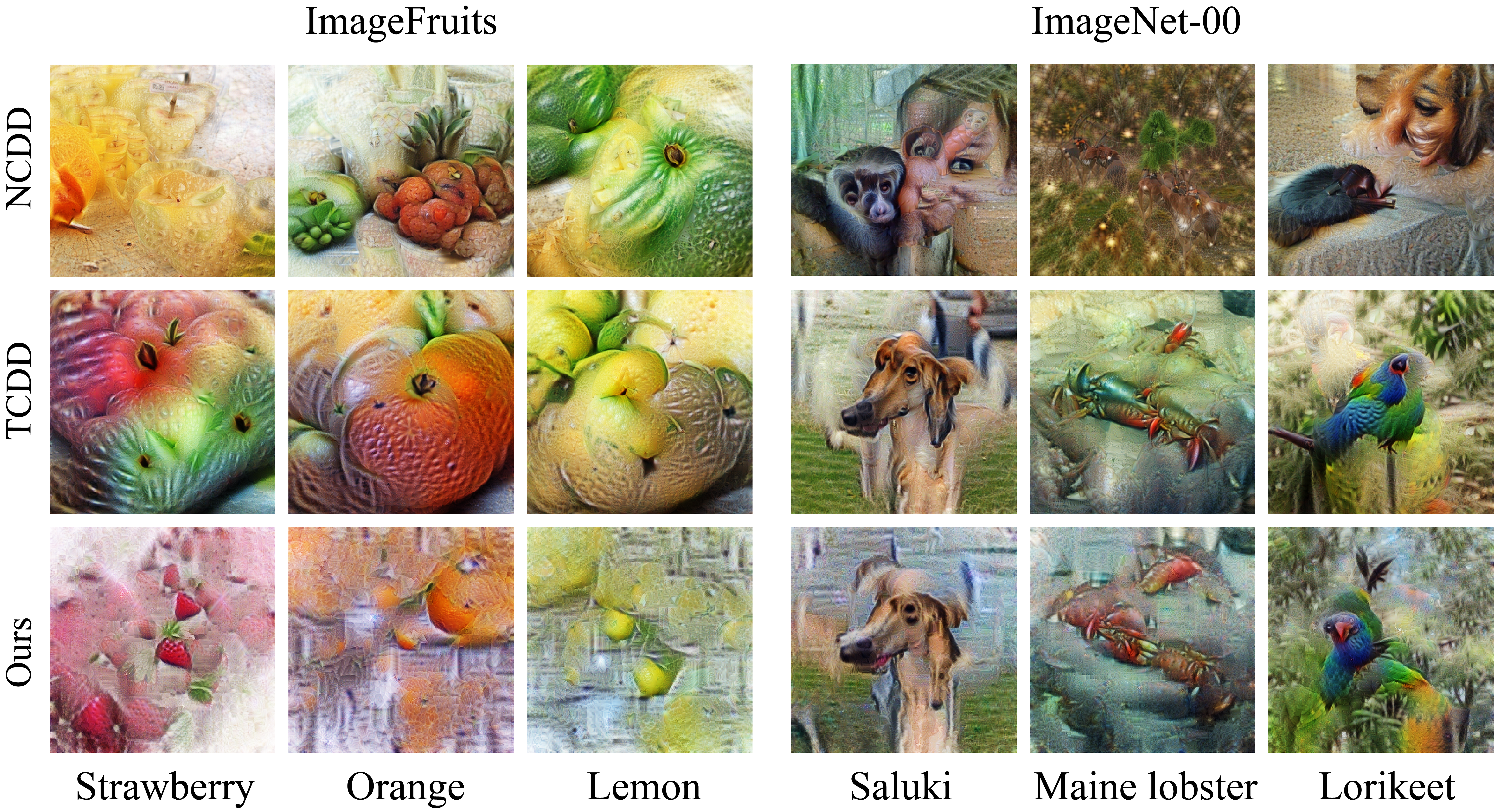}
        \vspace{-1.65em}
        \captionof{figure}{\looseness=-1 A comparison between images synthesized on ImageNet-100 by TCDD and NCDD. 
        }
        \vspace{-1em}
        \label{fig:tcdd_ncdd}
    \end{minipage}
\end{figure}

%% file: tables/extension.tex
\begin{table}[t]
  \centering
  \begin{minipage}[t]{0.49\textwidth}\setlength{\tabcolsep}{4.8pt}
    \centering\smaller             
    \renewcommand{\arraystretch}{1.25}
    \caption{Comparison of LGM and SFM under different IPC settings. The distillation model is DINO-v2 and the eval model is EVA-02.}
    \vspace{6pt}
    \label{tab:ipc}
    \begin{tabular}{lcccc}
    \hline
    IPC    & 1        & 2        & 5        & 10       \\ \hline
    Random & 64.5\smaller{\gray{$\pm$2.7}} & 72.9\smaller{\gray{$\pm$0.2}} & 80.9\smaller{\gray{$\pm$0.1}} & 85.0\smaller{\gray{$\pm$0.2}} \\
    MGD${}^{\text{3}}$ & 77.0\smaller{\gray{$\pm$0.2}} & 80.9\smaller{\gray{$\pm$0.0}} & 83.5\smaller{\gray{$\pm$0.1}} & 85.8\smaller{\gray{$\pm$0.1}} \\
    LGM  & 82.4\smaller{\gray{$\pm$0.2}} & 84.9\smaller{\gray{$\pm$0.2}} & 85.5\smaller{\gray{$\pm$0.0}} & 85.5\smaller{\gray{$\pm$0.1}} \\
    \textbf{SFM}  & \textbf{86.2}\smaller{\gray{$\pm$0.1}} & \textbf{86.5}\smaller{\gray{$\pm$0.1}} & \textbf{87.5}\smaller{\gray{$\pm$0.1}} & \textbf{88.4}\smaller{\gray{$\pm$0.0}} \\ \hline
    \end{tabular}
  \end{minipage}
  \hfill
  \begin{minipage}[t]{0.49\textwidth}\setlength{\tabcolsep}{3.6pt}
        \centering\smaller                 
    \renewcommand{\arraystretch}{1.1}
    \caption{Performance of our CI combined with existing distillation methods.}
    \vspace{2pt}
    \label{tab:ci}
    \begin{tabular}{lcccc}
    \hline
    
    Method     & CLIP                              & DINO-v2                             & EVA-02                               & MoCo-v3   \\  \hline
    MGD${}^{\text{3}}$       & 64.4\smaller{\gray{$\pm$0.4}}     & 84.0\smaller{\gray{$\pm$0.3}}       & 77.0\smaller{\gray{$\pm$0.2}}        & 71.9\smaller{\gray{$\pm$0.1}}        \\ 
    MGD${}^{\text{3}}$+\textbf{CI}      & 71.4\smaller{\gray{$\pm$0.1}}     & 94.8\smaller{\gray{$\pm$0.0}}       & 82.6\smaller{\gray{$\pm$0.1}}        & 74.6\smaller{\gray{$\pm$0.1}}        \\  \hline
    LGM      & 70.8\smaller{\gray{$\pm$0.1}}     & 89.8\smaller{\gray{$\pm$0.0}}       & 82.4\smaller{\gray{$\pm$0.2}}        & 72.5\smaller{\gray{$\pm$0.1}}        \\
    LGM+\textbf{CI}       & 76.7\smaller{\gray{$\pm$0.1}}     & 95.0\smaller{\gray{$\pm$0.0}}       & 84.6\smaller{\gray{$\pm$0.1}}        & 75.1\smaller{\gray{$\pm$0.0}}        \\ \hline
    SFM      & 78.6\smaller{\gray{$\pm$0.0}}     & 90.6\smaller{\gray{$\pm$0.0}}       & 86.2\smaller{\gray{$\pm$0.1}}        & 80.5\smaller{\gray{$\pm$0.0}}        \\
    SFM+\textbf{CI}       & \textbf{82.1\smaller{\gray{$\pm$0.0}}}     & \textbf{95.1\smaller{\gray{$\pm$0.1}}}       & \textbf{88.8\smaller{\gray{$\pm$0.1}}}        & \textbf{ 80.6\smaller{\gray{$\pm$0.1}} }       \\ \hline
\end{tabular}
  \end{minipage}
\end{table}

%% file: figures/soft_label.tex
\begin{figure}[t]
    \centering
    \hspace{0.1cm} 
    \begin{subfigure}{0.23\textwidth}
        \centering
        \includegraphics[width=\linewidth]{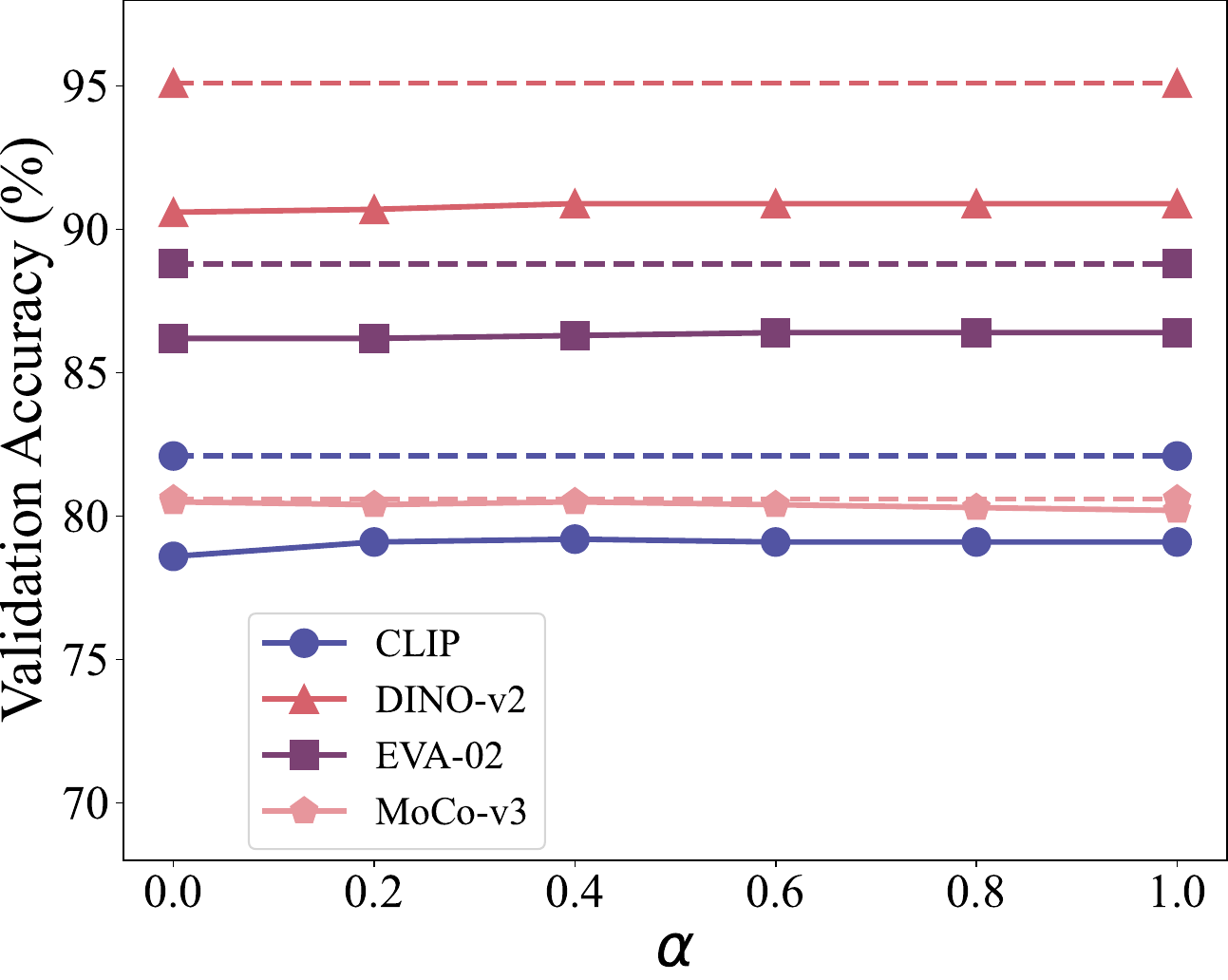}
        \caption{T=1}
    \end{subfigure}
    \hspace{0.1cm} 
    \begin{subfigure}{0.23\textwidth}
        \centering
        \includegraphics[width=\linewidth]{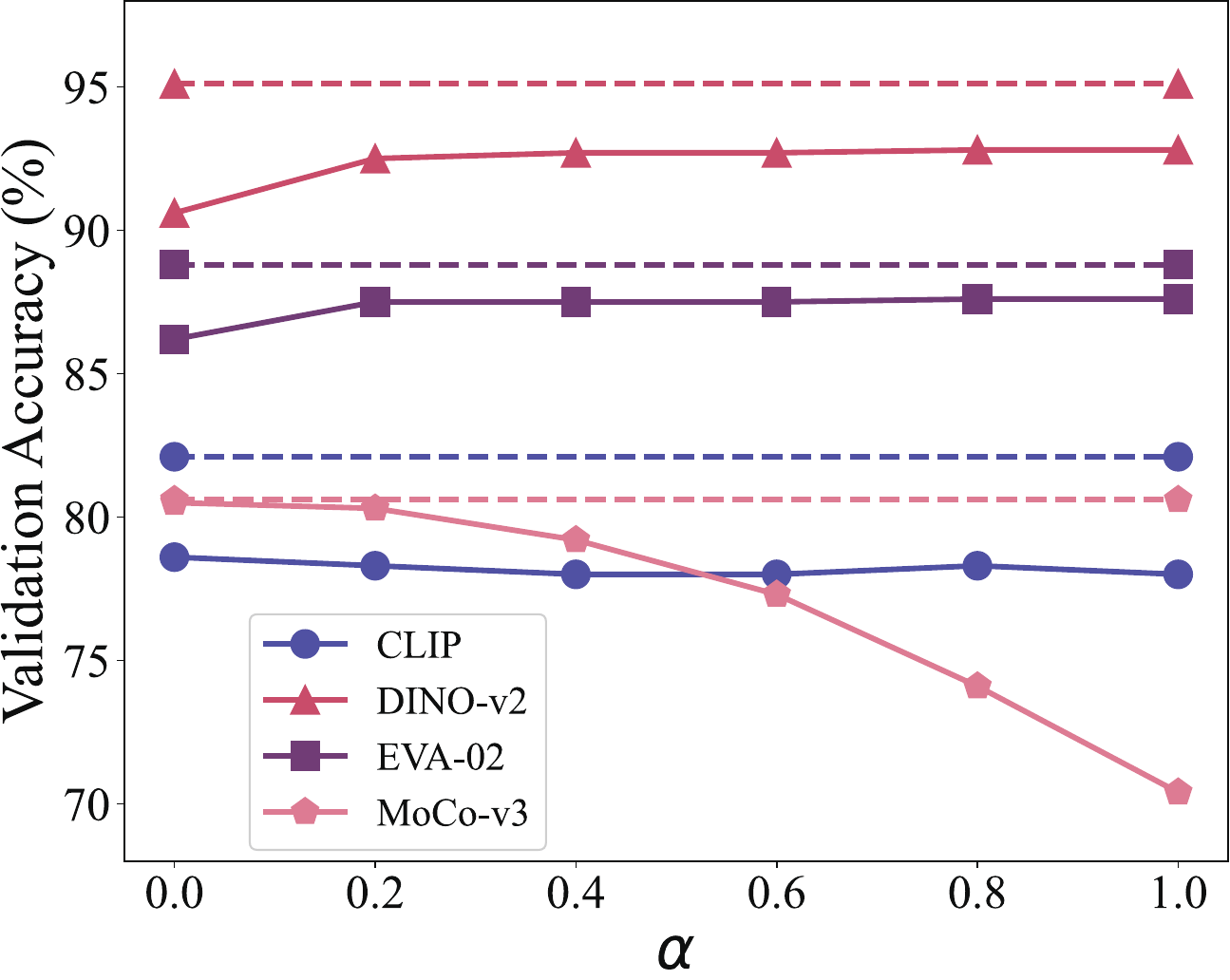}
        \caption{T=4}
    \end{subfigure}
    \hspace{0.1cm}
    \begin{subfigure}{0.23\textwidth}
        \centering
        \includegraphics[width=\linewidth]{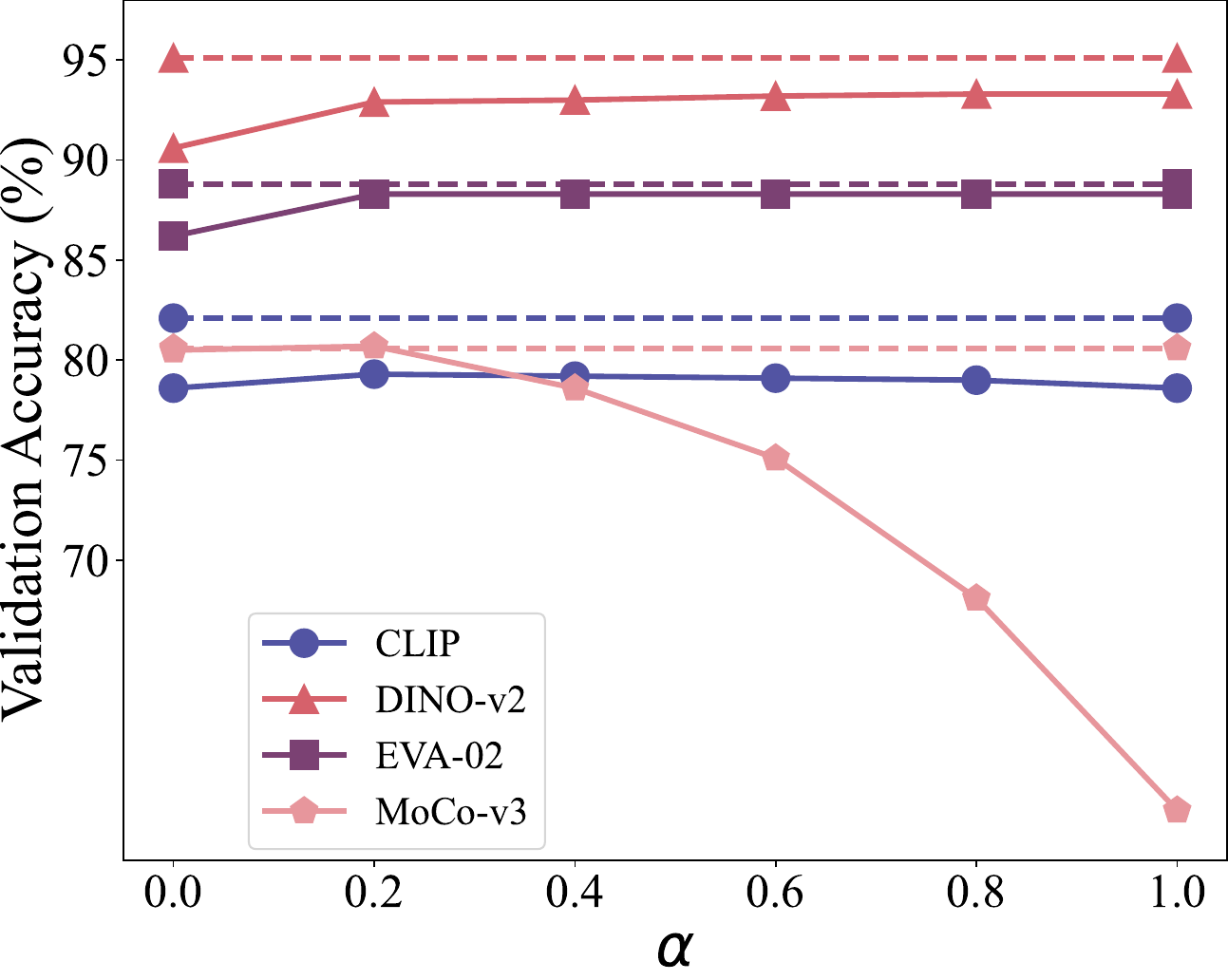}
        \caption{T=8}
    \end{subfigure}
    \hspace{0.1cm} 
    \begin{subfigure}{0.23\textwidth}
        \centering
        \includegraphics[width=\linewidth]{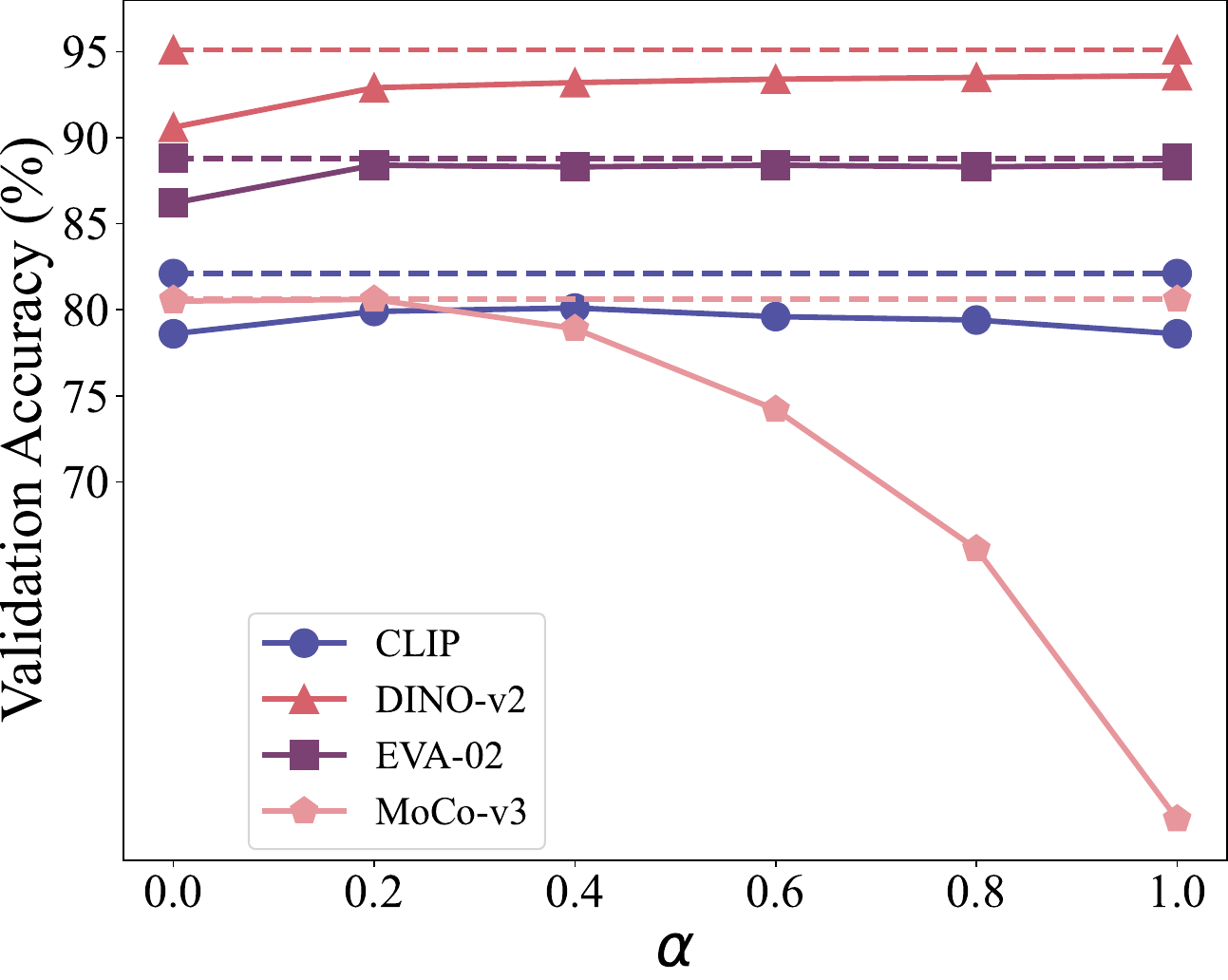}
        \caption{T=16}
    \end{subfigure}
    \caption{A comparison between soft label (solid
line) with different temperatures and our CI (dashed line). Our CI outperforms soft labels across all settings.}
    \label{fig:T}
    \vspace{-5pt}
\end{figure}

%% file: figures/syn_images.tex
\begin{figure*}[t]
    \centering
    \includegraphics[width=\textwidth]{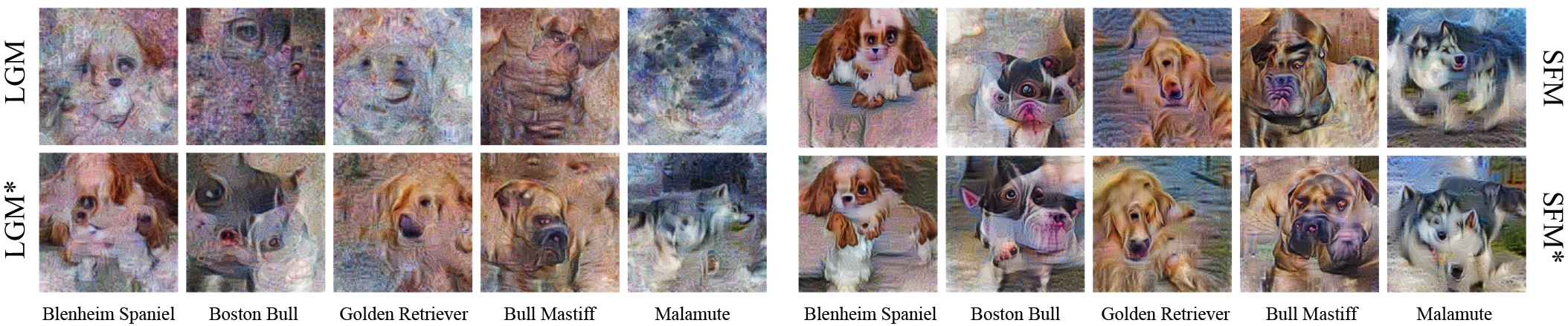}
    \caption{
    \looseness=-1
    A comparison of images synthesized by LGM versus our SFM under a single augmentation, alongside LGM* and SFM* under multiple augmentations. The dataset is Stanford Dogs.}
    \vspace{-6pt}
    \label{fig:syn_images}
\end{figure*}

 

%% file: figures/flow_compare.tex
\begin{figure}[t]
\vspace{-0pt}
    \centering
    \begin{subfigure}{0.31\textwidth}
        \centering
        \includegraphics[width=\linewidth]{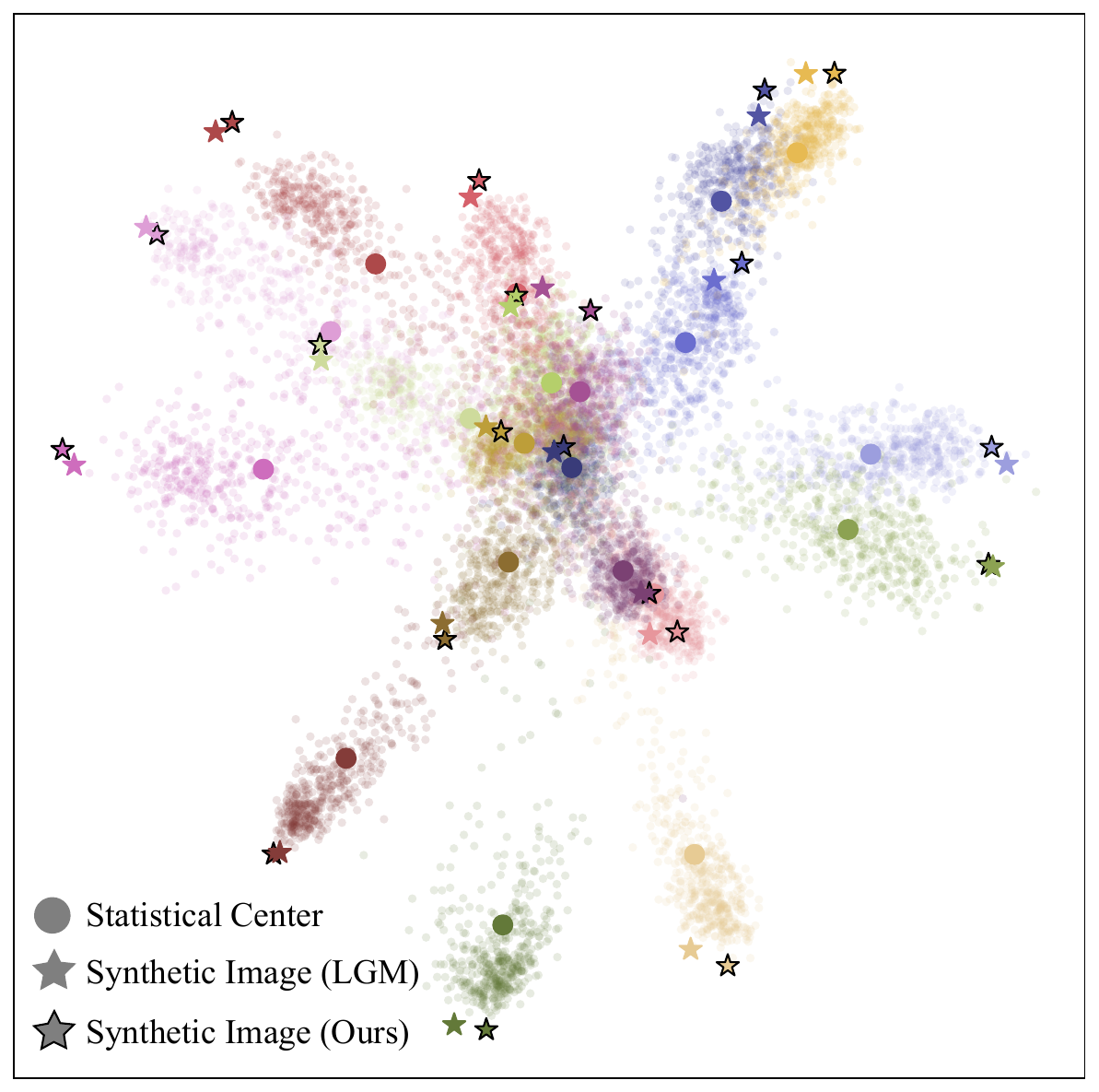}
        \caption{Distribution}
    \label{fig:lgm_flow}
    \end{subfigure}
    \hspace{0.3cm}
    \begin{subfigure}{0.31\textwidth}
        \centering
        \includegraphics[width=\linewidth]{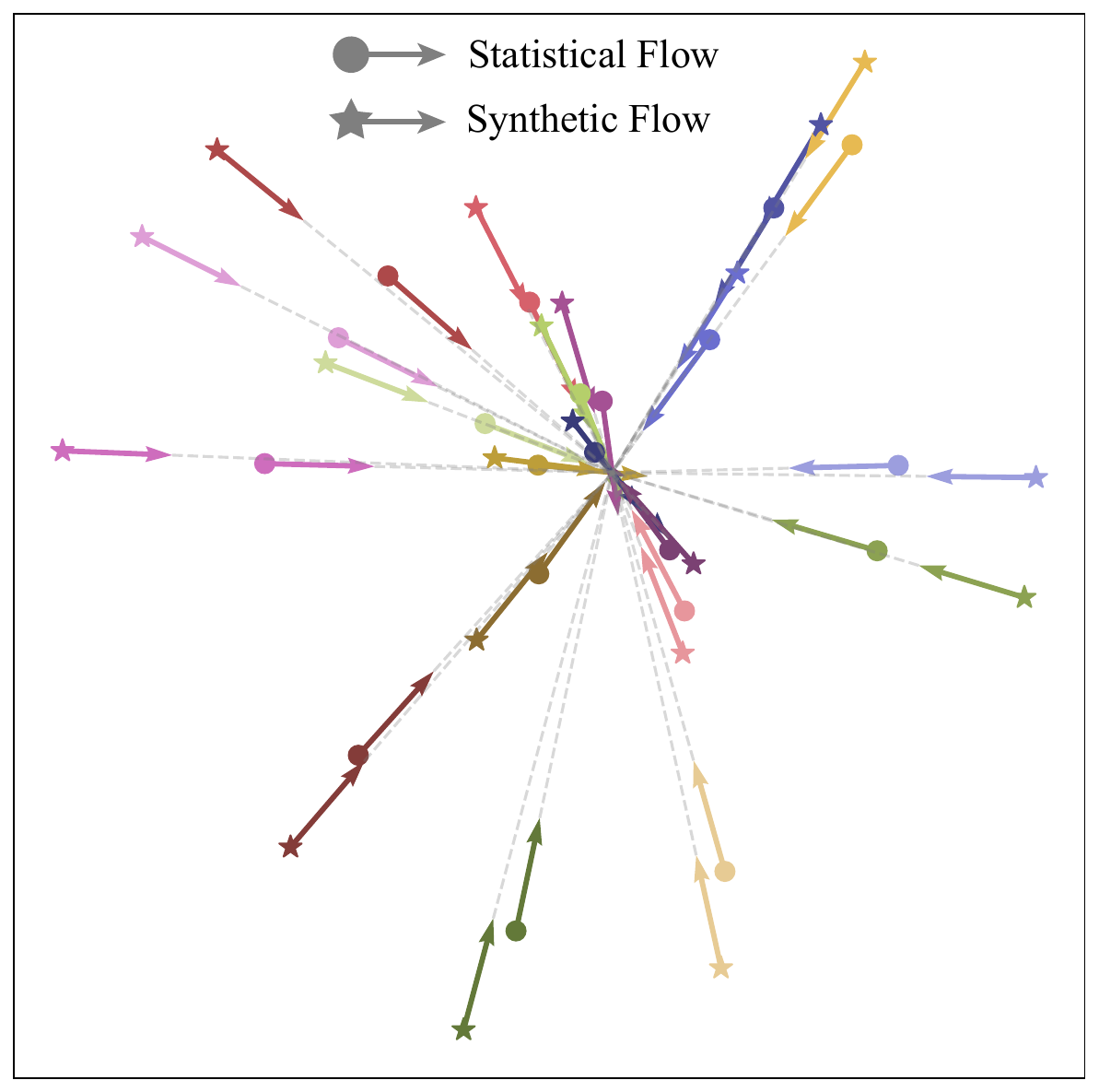}
        \caption{LGM}
        \label{fig:lgm_flow}
    \end{subfigure}
    \hspace{0.3cm} 
    \begin{subfigure}{0.31\textwidth}
        \centering
        \includegraphics[width=\linewidth]{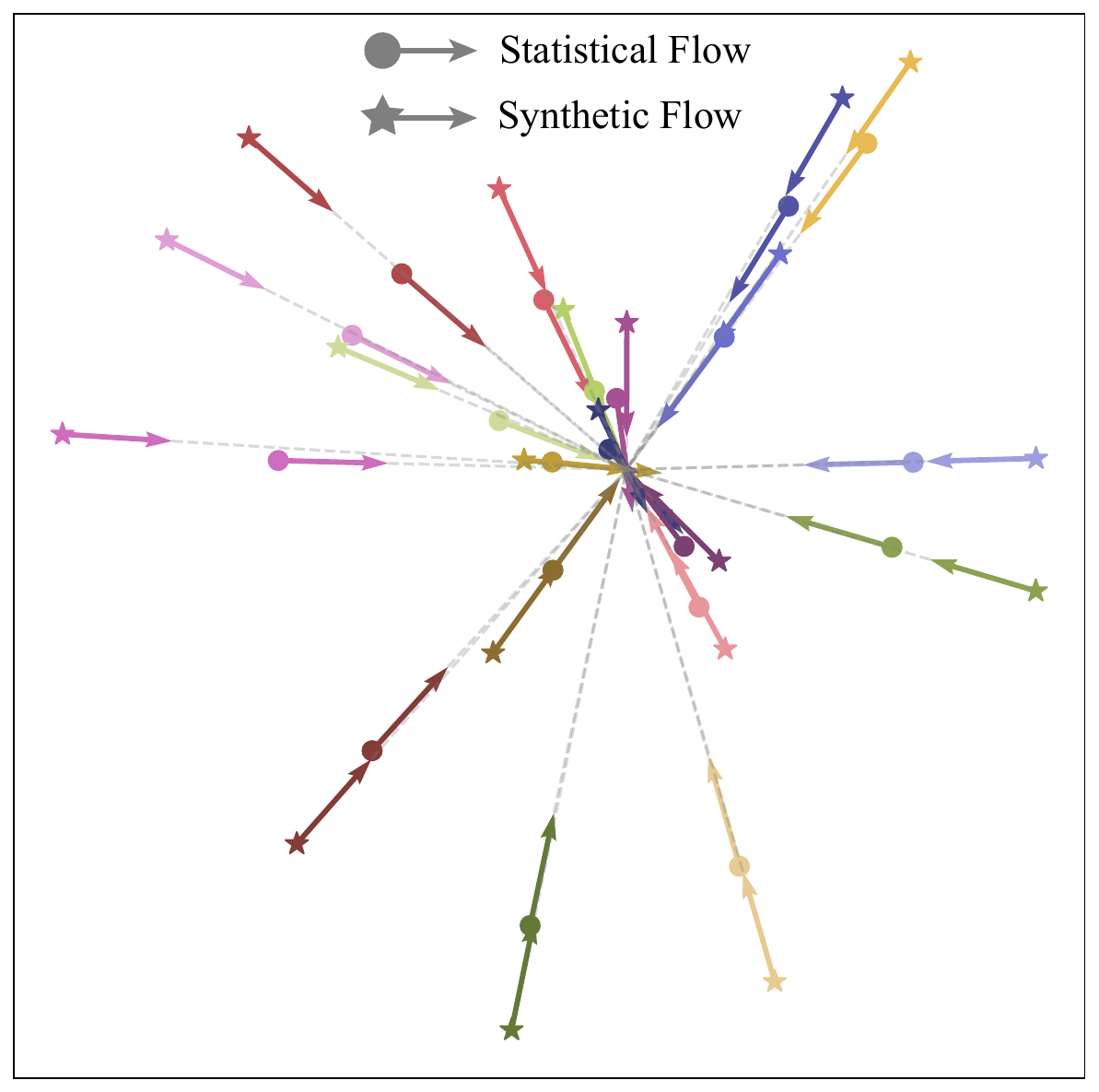}
        \caption{Ours}
        \label{fig:our_flow}
    \end{subfigure}
    \caption{The (a) distribution, along with the flows of (b) LGM  and (c) our SFM, on ImageNet-100. We perform PCA for dimensionality reduction, visualizing 20 classes. Both LGM and our SFM synthesize samples on the distribution boundary with discriminative attributes.}
    \vspace{-6pt}
    \label{fig:flow_both}
\end{figure}


%% file: tables/mean.tex
\begin{table}[h]
\small
\centering
\setlength{\tabcolsep}{3.5pt}
\renewcommand{\arraystretch}{1.1}
\caption{Effect of different gaussian noise means on performance. The distillation model is EVA-02. }
\vspace{4pt}
\label{tab:mean}
    \begin{tabular}{lcccc}
    \hline
    
    Mean     & CLIP                              & DINO-v2                             & EVA-02                               & MoCo-v3   \\  \hline
    0.5            & 81.0\smaller{\gray{$\pm$0.2}}     & 88.0\smaller{\gray{$\pm$0.2}}       & 88.9\smaller{\gray{$\pm$0.2}}        & 70.5\smaller{\gray{$\pm$0.0}}             \\
    0             & 82.2\smaller{\gray{$\pm$0.0}}     & 88.8\smaller{\gray{$\pm$0.0}}       & 88.6\smaller{\gray{$\pm$0.3}}        & 76.6\smaller{\gray{$\pm$0.0}}         \\\hline
\end{tabular}
\end{table}

%% file: tables/ci_train.tex
\begin{table*}[t]\small\setlength{\tabcolsep}{0pt}
	\caption{The performance of the gold classifier trained on various full datasets.}
	\begin{center}{
		\begin{tabularx}{1.0\textwidth}{lYYYYYYYYYY}
			\toprule\multirow{2}{*}{\shortstack{Epoch}} & \multicolumn{ 5}{c}{\normalsize ImageNet-100} & \multicolumn{ 5}{c}{\normalsize ImageNet-1k}\\\cmidrule(lr){2-6}\cmidrule(lr){7-11}
			& {CLIP} & {DINO-v2} & {EVA-02} & {MoCo-v3} & {Average} & {CLIP} & {DINO-v2} & {EVA-02} & {MoCo-v3} & {Average} \\ \midrule
            10	& 92.8\smaller{\gray{$\pm$0.1}}	& 95.3\smaller{\gray{$\pm$0.1}}	& 94.5\smaller{\gray{$\pm$0.0}}	& 88.2\smaller{\gray{$\pm$0.1}}	& 92.7\smaller{\gray{$\pm$0.1}}	& 79.0\smaller{\gray{$\pm$0.0}}	& 82.0\smaller{\gray{$\pm$0.0}}	& 82.4\smaller{\gray{$\pm$0.1}}	& 73.9\smaller{\gray{$\pm$0.0}}	& 79.3\smaller{\gray{$\pm$0.0}}	\\ \midrule 
           \multirow{2}{*}{\shortstack{Epoch}} & \multicolumn{5}{c}{\normalsize Stanford Dogs} & \multicolumn{5}{c}{\normalsize CUB-200}\\\cmidrule(lr){2-6}\cmidrule(lr){7-11}
			& {CLIP} & {DINO-v2} & {EVA-02} & {MoCo-v3} & {Average} & {CLIP} & {DINO-v2} & {EVA-02} & {MoCo-v3} & {Average} \\ \midrule
            10	& 72.0\smaller{\gray{$\pm$0.1}}	& 88.8\smaller{\gray{$\pm$0.0}}	& 83.2\smaller{\gray{$\pm$0.1}}	& 70.4\smaller{\gray{$\pm$0.3}}	& 78.6\smaller{\gray{$\pm$0.1}}	& 72.5\smaller{\gray{$\pm$0.4}}	& 89.7\smaller{\gray{$\pm$0.1}}	& 83.2\smaller{\gray{$\pm$0.1}}	& 45.3\smaller{\gray{$\pm$0.4}}	& 72.7\smaller{\gray{$\pm$0.3}}	\\  \midrule 
		\end{tabularx}
	}\end{center} 
	\label{tab:gc_acc}
\end{table*}

%% file: tables/cross1k.tex
\begin{table*}\smaller\setlength{\tabcolsep}{0pt}
	\caption{The cross-model generalization performance on ImageNet-1k.}
    \vspace{-8pt}
	\begin{center}{
		\begin{tabularx}{1.0\textwidth}{lYYYYYYYYYY}
        \toprule
			\shortstack{Distill. Model} & \multicolumn{ 5}{c}{\normalsize CLIP} & \multicolumn{ 5}{c}{\normalsize DINO-v2}\\\cmidrule(lr){2-6}\cmidrule(lr){7-11}
			\shortstack{Eval. Model}& {CLIP} & {DINO-v2} & {EVA-02} & {MoCo-v3} & {Average} & {CLIP} & {DINO-v2} & {EVA-02} & {MoCo-v3} & {Average} \\ \midrule
            LGM*	& \gray{63.0}\smaller{\gray{$\pm$0.0}}	& 56.4\smaller{\gray{$\pm$0.1}}	& 59.7\smaller{\gray{$\pm$0.1}}	& 39.5\smaller{\gray{$\pm$0.0}}	& 54.7\smaller{\gray{$\pm$0.1}}	& 54.1\smaller{\gray{$\pm$0.0}}	& \gray{75.0}\smaller{\gray{$\pm$0.1}}	& 65.4\smaller{\gray{$\pm$0.1}}	& \textbf{60.0\smaller{\gray{$\pm$0.0}}}	& 63.7\smaller{\gray{$\pm$0.1}}	\\ 
            LGM	& \gray{55.5}\smaller{\gray{$\pm$0.0}}	& 45.0\smaller{\gray{$\pm$0.0}}	& 48.5\smaller{\gray{$\pm$0.0}}	& 25.8\smaller{\gray{$\pm$0.0}}	& 43.7\smaller{\gray{$\pm$0.0}}	& 46.7\smaller{\gray{$\pm$0.0}}	& \gray{73.5}\smaller{\gray{$\pm$0.2}}	& 61.2\smaller{\gray{$\pm$0.1}}	& 54.7\smaller{\gray{$\pm$0.1}}	& 59.0\smaller{\gray{$\pm$0.1}}	\\ 
            \textbf{SFM}	& \textbf{\gray{67.1}\smaller{\gray{$\pm$0.0}}}	& \textbf{64.2\smaller{\gray{$\pm$0.1}}}	& \textbf{64.7\smaller{\gray{$\pm$0.0}}}	& \textbf{46.6\smaller{\gray{$\pm$0.0}}}	& \textbf{60.7\smaller{\gray{$\pm$0.1}}}	& \textbf{56.6\smaller{\gray{$\pm$0.0}}}	& \textbf{\gray{75.1}\smaller{\gray{$\pm$0.1}}}	& \textbf{65.6\smaller{\gray{$\pm$0.0}}}	& 59.9\smaller{\gray{$\pm$0.0}}	& \textbf{64.3\smaller{\gray{$\pm$0.0}}}	\\ \midrule
			\shortstack{Distill. Model} & \multicolumn{ 5}{c}{\normalsize EVA-02} & \multicolumn{ 5}{c}{\normalsize MoCo-v3}\\\cmidrule(lr){2-6}\cmidrule(lr){7-11}
			\shortstack{Eval. Model}& {CLIP} & {DINO-v2} & {EVA-02} & {MoCo-v3} & {Average} & {CLIP} & {DINO-v2} & {EVA-02} & {MoCo-v3} & {Average} \\ \midrule
            LGM*	& 55.5\smaller{\gray{$\pm$0.1}}	& 65.9\smaller{\gray{$\pm$0.1}}	& \gray{70.3}\smaller{\gray{$\pm$0.1}}	& \textbf{51.8\smaller{\gray{$\pm$0.0}}}	& 60.9\smaller{\gray{$\pm$0.1}}	& 41.4\smaller{\gray{$\pm$0.0}}	& \textbf{66.9\smaller{\gray{$\pm$0.1}}}	& 57.2\smaller{\gray{$\pm$0.1}}	& \gray{63.2}\smaller{\gray{$\pm$0.0}}	& 57.2\smaller{\gray{$\pm$0.1}}	\\ 
            LGM	& 47.8\smaller{\gray{$\pm$0.1}}	& 61.1\smaller{\gray{$\pm$0.1}}	& \gray{67.6}\smaller{\gray{$\pm$0.0}}	& 38.9\smaller{\gray{$\pm$0.0}}	& 53.9\smaller{\gray{$\pm$0.1}}	& 32.9\smaller{\gray{$\pm$0.0}}	& 61.5\smaller{\gray{$\pm$0.1}}	& 47.5\smaller{\gray{$\pm$0.0}}	& \gray{61.6}\smaller{\gray{$\pm$0.0}}	& 50.9\smaller{\gray{$\pm$0.1}}	\\ 
            \textbf{SFM}	& \textbf{58.8\smaller{\gray{$\pm$0.1}}}	& \textbf{70.0\smaller{\gray{$\pm$0.2}}}	& \textbf{\gray{71.6}\smaller{\gray{$\pm$0.1}}}	& 51.2\smaller{\gray{$\pm$0.0}}	& \textbf{62.9\smaller{\gray{$\pm$0.1}}}	& \textbf{47.0\smaller{\gray{$\pm$0.1}}}	& 65.2\smaller{\gray{$\pm$0.1}}	& \textbf{57.5\smaller{\gray{$\pm$0.1}}}	& \textbf{\gray{63.4}\smaller{\gray{$\pm$0.0}}}	& \textbf{58.3\smaller{\gray{$\pm$0.1}}}	\\ \midrule 
			Full Dataset	& 78.7\smaller{\color{darkgray}{$\pm$0.0}}	& 83.0\smaller{\color{darkgray}{$\pm$0.0}}	& 81.7\smaller{\color{darkgray}{$\pm$0.1}}	& 76.5\smaller{\color{darkgray}{$\pm$0.0}}	& 80.0\smaller{\color{darkgray}{$\pm$0.0}}  & 78.7\smaller{\color{darkgray}{$\pm$0.0}}	& 83.0\smaller{\color{darkgray}{$\pm$0.0}}	& 81.7\smaller{\color{darkgray}{$\pm$0.1}}	& 76.5\smaller{\color{darkgray}{$\pm$0.0}}	& 80.0\smaller{\color{darkgray}{$\pm$0.0}}	\\\bottomrule
		\end{tabularx}
	}\end{center} 
	\label{tab:cross1k}
\end{table*}

%% file: tables/crosscub.tex
\begin{table}\smaller\setlength{\tabcolsep}{0pt}
	\vspace{-8pt}
    \caption{The cross-model generalization performance on CUB-200.}
	\begin{center}{
		\begin{tabularx}{1.0\textwidth}{lYYYYYYYYYY}
        \toprule
			\shortstack{Distill. Model} & \multicolumn{ 5}{c}{\normalsize CLIP} & \multicolumn{ 5}{c}{\normalsize DINO-v2}\\\cmidrule(lr){2-6}\cmidrule(lr){7-11}
			\shortstack{Eval. Model}& {CLIP} & {DINO-v2} & {EVA-02} & {MoCo-v3} & {Average} & {CLIP} & {DINO-v2} & {EVA-02} & {MoCo-v3} & {Average} \\ \midrule
            LGM	& \gray{16.4}\smaller{\gray{$\pm$0.0}}	& 15.2\smaller{\gray{$\pm$0.0}}	& 13.6\smaller{\gray{$\pm$0.0}}	& 2.7\smaller{\gray{$\pm$0.1}}	& 12.0\smaller{\gray{$\pm$0.0}}	& 22.1\smaller{\gray{$\pm$0.2}}	& \gray{81.1}\smaller{\gray{$\pm$0.1}}	& 50.3\smaller{\gray{$\pm$0.1}}	& 26.0\smaller{\gray{$\pm$0.1}}	& 44.9\smaller{\gray{$\pm$0.1}}	\\ 
            \textbf{SFM}	& \textbf{\gray{67.6}\smaller{\gray{$\pm$0.2}}}	& \textbf{51.4\smaller{\gray{$\pm$0.3}}}	& \textbf{48.3\smaller{\gray{$\pm$0.3}}}	& \textbf{16.2\smaller{\gray{$\pm$0.3}}}	& \textbf{45.9\smaller{\gray{$\pm$0.3}}}	& \textbf{37.8\smaller{\gray{$\pm$0.1}}}	& \textbf{\gray{85.7}\smaller{\gray{$\pm$0.1}}}	& \textbf{61.7\smaller{\gray{$\pm$0.1}}}	& \textbf{29.5\smaller{\gray{$\pm$0.1}}}	& \textbf{53.7\smaller{\gray{$\pm$0.1}}}	\\ \midrule
			\shortstack{Distill. Model} & \multicolumn{ 5}{c}{\normalsize EVA-02} & \multicolumn{ 5}{c}{\normalsize MoCo-v3}\\\cmidrule(lr){2-6}\cmidrule(lr){7-11}
			\shortstack{Eval. Model}& {CLIP} & {DINO-v2} & {EVA-02} & {MoCo-v3} & {Average} & {CLIP} & {DINO-v2} & {EVA-02} & {MoCo-v3} & {Average} \\ \midrule
            LGM	& 17.8\smaller{\gray{$\pm$0.1}}	& 37.4\smaller{\gray{$\pm$0.1}}	& \gray{51.5}\smaller{\gray{$\pm$0.2}}	& 8.9\smaller{\gray{$\pm$0.1}}	& 28.9\smaller{\gray{$\pm$0.1}}	& 6.4\smaller{\gray{$\pm$0.0}}	& 36.4\smaller{\gray{$\pm$0.5}}	& 17.5\smaller{\gray{$\pm$0.3}}	& \gray{25.5}\smaller{\gray{$\pm$0.0}}	& 21.5\smaller{\gray{$\pm$0.2}}	\\ 
            \textbf{SFM}	& \textbf{43.4\smaller{\gray{$\pm$0.0}}}	& \textbf{66.8\smaller{\gray{$\pm$0.3}}}	& \textbf{\gray{72.5}\smaller{\gray{$\pm$0.2}}}	& 18.3\smaller{\gray{$\pm$0.1}}	& \textbf{50.3\smaller{\gray{$\pm$0.2}}}	& \textbf{14.8\smaller{\gray{$\pm$0.1}}}	& \textbf{51.7\smaller{\gray{$\pm$0.0}}}	& \textbf{33.1\smaller{\gray{$\pm$0.0}}}	& \textbf{\gray{41.4}\smaller{\gray{$\pm$0.1}}}	& \textbf{35.3\smaller{\gray{$\pm$0.1}}}	\\ \midrule 
			Full Dataset	& 77.5\smaller{\gray{$\pm$0.7}}	& 90.2\smaller{\gray{$\pm$0.2}}	& 84.0\smaller{\gray{$\pm$0.3}}	& 43.7\smaller{\gray{$\pm$0.8}}	& 73.8\smaller{\gray{$\pm$0.5}}   &
            77.5\smaller{\gray{$\pm$0.7}}	& 90.2\smaller{\gray{$\pm$0.2}}	& 84.0\smaller{\gray{$\pm$0.3}}	& 43.7\smaller{\gray{$\pm$0.8}}	& 73.8\smaller{\gray{$\pm$0.5}}	\\\bottomrule
		\end{tabularx}
	}\end{center} 
	\label{tab:crosscub}
\end{table}

%% file: tables/crossdog.tex
\begin{table}\smaller\setlength{\tabcolsep}{0pt}
    \vspace{{-8pt}}
    \caption{The cross-model generalization performance on Stanford Dogs.}
	\begin{center}{
		\begin{tabularx}{1.0\textwidth}{lYYYYYYYYYY}
        \toprule
			\shortstack{Distill. Model} & \multicolumn{ 5}{c}{\normalsize CLIP} & \multicolumn{ 5}{c}{\normalsize DINO-v2}\\\cmidrule(lr){2-6}\cmidrule(lr){7-11}
			\shortstack{Eval. Model}& {CLIP} & {DINO-v2} & {EVA-02} & {MoCo-v3} & {Average} & {CLIP} & {DINO-v2} & {EVA-02} & {MoCo-v3} & {Average} \\ \midrule
            LGM	& \gray{15.0}\smaller{\gray{$\pm$0.0}}	& 17.7\smaller{\gray{$\pm$0.2}}	& 16.3\smaller{\gray{$\pm$0.0}}	& 6.2\smaller{\gray{$\pm$0.1}}	& 13.8\smaller{\gray{$\pm$0.1}}	& 27.5\smaller{\gray{$\pm$0.1}}	& \gray{79.8}\smaller{\gray{$\pm$0.2}}	& 59.7\smaller{\gray{$\pm$0.0}}	& 56.7\smaller{\gray{$\pm$0.6}}	& 55.9\smaller{\gray{$\pm$0.2}}	\\ 
            \textbf{SFM}	& \textbf{\gray{55.0}\smaller{\gray{$\pm$0.3}}}	& \textbf{62.6\smaller{\gray{$\pm$0.7}}}	& \textbf{60.0\smaller{\gray{$\pm$0.3}}}	& \textbf{45.6\smaller{\gray{$\pm$0.8}}}	& \textbf{55.8\smaller{\gray{$\pm$0.5}}}	& \textbf{39.2\smaller{\gray{$\pm$0.2}}}	& \textbf{\gray{81.8}\smaller{\gray{$\pm$0.2}}}	& \textbf{66.0\smaller{\gray{$\pm$0.4}}}	& \textbf{62.3\smaller{\gray{$\pm$0.2}}}	& \textbf{62.3\smaller{\gray{$\pm$0.3}}}	\\ \midrule
			\shortstack{Distill. Model} & \multicolumn{ 5}{c}{\normalsize EVA-02} & \multicolumn{ 5}{c}{\normalsize MoCo-v3}\\\cmidrule(lr){2-6}\cmidrule(lr){7-11}
			\shortstack{Eval. Model}& {CLIP} & {DINO-v2} & {EVA-02} & {MoCo-v3} & {Average} & {CLIP} & {DINO-v2} & {EVA-02} & {MoCo-v3} & {Average} \\ \midrule 
            LGM	& 16.0\smaller{\gray{$\pm$0.0}}	& 53.8\smaller{\gray{$\pm$0.3}}	& \gray{65.0}\smaller{\gray{$\pm$0.2}}	& 32.5\smaller{\gray{$\pm$0.2}}	& 41.8\smaller{\gray{$\pm$0.2}}	& 11.3\smaller{\gray{$\pm$0.3}}	& 66.1\smaller{\gray{$\pm$0.4}}	& 48.9\smaller{\gray{$\pm$0.1}}	& \gray{63.5}\smaller{\gray{$\pm$0.2}}	& 47.5\smaller{\gray{$\pm$0.3}}	\\ 
            \textbf{SFM}	& \textbf{38.6\smaller{\gray{$\pm$0.0}}}	& \textbf{72.7\smaller{\gray{$\pm$0.4}}}	& \textbf{\gray{72.9}\smaller{\gray{$\pm$0.2}}}	& \textbf{53.2\smaller{\gray{$\pm$0.1}}}	& \textbf{59.4\smaller{\gray{$\pm$0.2}}}	& \textbf{14.6\smaller{\gray{$\pm$0.3}}}	& 70.7\smaller{\gray{$\pm$0.3}}	& \textbf{57.4\smaller{\gray{$\pm$0.0}}}	& \textbf{\gray{67.2}\smaller{\gray{$\pm$0.3}}}	& \textbf{52.5\smaller{\gray{$\pm$0.2}}}	\\ \midrule 

            Full Dataset	& 76.9\smaller{\gray{$\pm$0.1}}	& 88.6\smaller{\gray{$\pm$0.1}}	& 82.6\smaller{\gray{$\pm$0.1}}	& 72.3\smaller{\gray{$\pm$0.5}}	& 80.1\smaller{\gray{$\pm$0.2}}   & 76.9\smaller{\gray{$\pm$0.1}}	& 88.6\smaller{\gray{$\pm$0.1}}	& 82.6\smaller{\gray{$\pm$0.1}}	& 72.3\smaller{\gray{$\pm$0.5}}	& 80.1\smaller{\gray{$\pm$0.2}}	\\\bottomrule
		\end{tabularx}
	}\end{center} 
	\label{tab:crossdog}
    \vspace{-18pt}
\end{table}

%% file: figures/style.tex
\begin{figure*}[t]
    \centering
    \includegraphics[width=\textwidth]{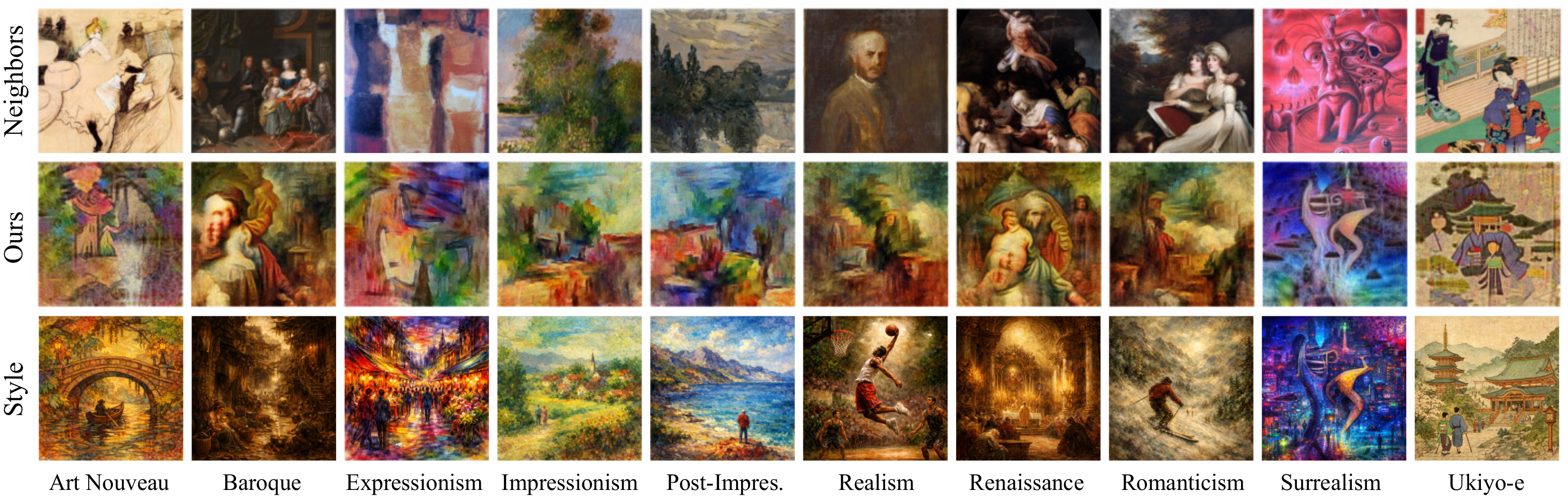}
    \caption{
    \looseness=-1
    Distilling the ArtBench \cite{liao2022artbench} dataset.
    }
    \label{fig:style}
\end{figure*}

%% file: tables/vitsbl.tex
\begin{table}[h!]\small\centering
\setlength{\tabcolsep}{4.8pt}
\renewcommand{\arraystretch}{1.2}
\caption{Performance of our SFM and CI under homogeneous architectures with different scales.}
\vspace{4pt}
\label{tab:vitsbl}
\begin{tabular}{lccc}
\hline
Size   & ViT-S(dim=384) & ViT-B(dim=768) & ViT-L(dim=1024) \\ \hline
SFM    & 84.9 \smaller{\gray{$\pm$0.3}}       & 90.6 \smaller{\gray{$\pm$0.0}}      & 90.4 \smaller{\gray{$\pm$0.3}}        \\
SFM+CI & 86.0 \smaller{\gray{$\pm$0.1}} & 95.1 \smaller{\gray{$\pm$0.1}} & 92.4  \smaller{\gray{$\pm$0.1}} \\ \hline
\end{tabular}
\end{table}

%% file: tables/gpu.tex
\begin{table}[h]\small\centering
\setlength{\tabcolsep}{4.8pt}
\renewcommand{\arraystretch}{1.2}
\caption{Quantitative comparison between LGM and our SFM in terms of distillation time and GPU memory usage. The distillation model is EVA-02.}
\vspace{4pt}
\label{tab:gpu}
\begin{tabular}{lcccccc}
\hline
\multicolumn{2}{l}{Augmentation} & 1    & 3    & 5    & 7     & 10    \\ \hline
LGM & \multirow{2}{*}{GPU(GB)}   & 16.6 & 48.5 & 84.5 & 107.3 & 165.2 \\
SFM &                            & 16.3 & 48.1 & 84.2 & 106.8 & 164.5 \\ \hline
LGM & \multirow{2}{*}{Time(min)} & 21   & 35   & 48   & 63    & 81    \\
SFM &                            & 18   & 28   & 38   & 48    & 62    \\ \hline
\end{tabular}
\end{table}